\documentclass[runningheads]{llncs}

\usepackage{eccv}

\usepackage{eccvabbrv}

\usepackage{graphicx}
\usepackage{booktabs}
\usepackage[accsupp]{axessibility}  % Improves PDF readability for those with disabilities.
\usepackage{algorithm}
\usepackage{algorithmic}
\usepackage{multirow}
\usepackage{makecell}

\let\titleold\title
\renewcommand{\title}[1]{%
  \titleold{#1}%
  \newcommand{\thetitle}{#1}%
}
\def\maketitlesupplementary{%
  \newpage
  \begin{center}
    {\Large \textbf{\thetitle}\\[0.5em]
    {\textit{Supplementary Material}}}\\[1em]
  \end{center}
}
\usepackage{hyperref}

\usepackage{orcidlink}

\usepackage{caption}
\begin{document}

\title{Leveling3D: Leveling Up 3D Reconstruction with Feed-Forward 3D Gaussian Splatting and Geometry-Aware Generation} 

\titlerunning{Leveling3D}

% Second Author\inst{2,3}\orcidlink{1111-2222-3333-4444} \and
% Third Author\inst{3}\orcidlink{2222--3333-4444-5555}}

\author{Yiming Huang\inst{1}\thanks{equal contribution,  $^\dagger$corresponding author} \and
Baixiang Huang\inst{1}$^{\star}$ \and
Beilei Cui\inst{1} \and
CHI KIT NG\inst{1} \and \\
Long Bai\inst{1} \and
Hongliang Ren\inst{1}$^\dagger$
}

\authorrunning{Y.~Huang et al.}

\institute{The Chinese University of Hong Kong\\
\email{\{yhuangdl\}@link.cuhk.edu.hk}}

\maketitle

\begin{center}
    \centering
    % \fbox{\rule{0pt}{3in} \rule{\linewidth}{0pt}}
    \includegraphics[width=\textwidth]{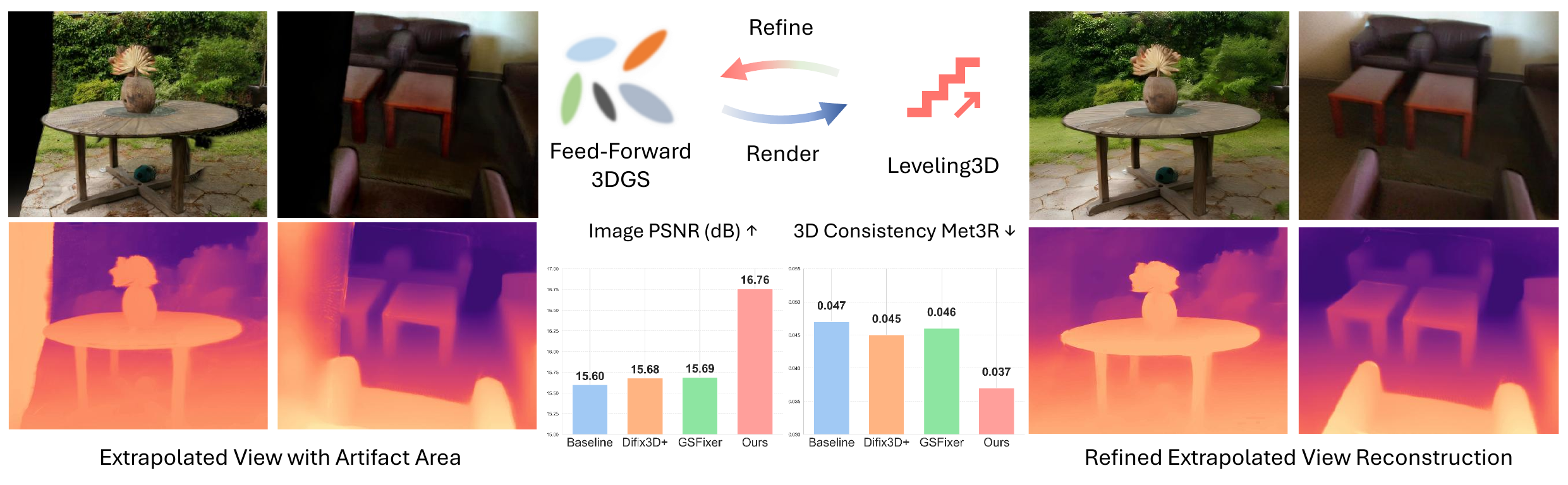}
    \captionof{figure}{We propose Leveling3D, a new method for extrapolated view refinement for Feed-Forward 3DGS with sparse input. Previous 3DGS refinement methods with naive reference or text prompts as diffusion control have limited geometry generalization ability within the artifact area in extrapolated views. In contrast, our method utilizes geometry-prior control, which generates fine RGB details with geometric consistency. The refined views further level up the 3D reconstruction with extrapolation to the unseen area, achieving SOTA performance on both image and depth synthesis.
    }
    \label{fig:teaser}
\end{center}%

\begin{abstract}
Feed-forward 3D reconstruction has revolutionized 3D vision, providing a powerful baseline for downstream tasks such as novel-view synthesis with 3D Gaussian Splatting. Previous works explore fixing the corrupted rendering results with a diffusion model. However, they lack geometric concern and fail at filling the missing area on the extrapolated view. In this work, we introduce Leveling3D, a novel pipeline that integrates feed-forward 3D reconstruction with geometrical-consistent generation to enable holistic simultaneous reconstruction and generation. We propose a geometry-aware leveling adapter, a lightweight technique that aligns internal knowledge in the diffusion model with the geometry prior from the feed-forward model. The leveling adapter enables generation on the artifact area of the extrapolated novel views caused by underconstrained regions of the 3D representation. Specifically, to learn a more diverse distributed generation, we introduce the palette filtering strategy for training, and a test-time masking refinement to prevent messy boundaries along the fixing regions. More importantly, the enhanced extrapolated novel views from Leveling3D could be used as the inputs for feed-forward 3DGS, leveling up the 3D reconstruction. We achieve SOTA performance on public datasets, including tasks such as novel-view synthesis and depth estimation.
\end{abstract}    
\section{Introduction}
\label{sec:intro}
3D reconstruction stands as a cornerstone technology for augmented/virtual reality, autonomous systems, and digital twin applications. Traditional pipelines~\cite{schonberger2016structure, mur2015orb} relying on Structure-from-Motion (SfM) and Multi-View Stereo (MVS) achieve remarkable precision through iterative optimization, but still suffer from computational complexity and fragile performance in textureless regions. The recent paradigm shift toward feed-forward approaches~\cite{dust3r_cvpr24, leroy2024mast3r, wang2025vggt, keetha2025mapanything, wang2025pi3} has revolutionized this landscape, enabling real-time, generalizable 3D reconstruction by training forward-pass deep networks with a large amount of data. The powerful baselines like MASt3R~\cite{leroy2024mast3r} and VGGT~\cite{wang2025vggt} have demonstrated unprecedented capability in joint camera poses and dense geometry prediction, establishing a strong foundation with robust geometry prior for downstream tasks, especially novel view synthesis (NVS)~\cite{wang2025volsplat, jiang2025anysplat, ye2024nopo} with 3D Gaussian Splatting (3DGS)~\cite{kerbl3Dgaussians}.

Despite recent advances, feed-forward 3DGS~\cite{wang2025volsplat, jiang2025anysplat, ye2024nopo} remains fundamentally limited for extrapolated view renderings when input views are sparse. Specifically, when rendering extrapolated views has large area outside the input boundaries, the rendered results from 3DGS exhibits severe artifacts such as incomplete geometry, texture corruption, and large blank areas where no observations exist as in \cref{fig:teaser}. Previous latent diffusion models~\cite{Rombach_2022_SD2, suvorov2021resolutionLaMa} excel at 2D inpainting, they lack knowledge of the underlying 3D structure, producing {multi-view inconsistent} hallucinations that violate scene geometry when projected across viewpoints. While, existing works~\cite{wu2025difix3d+,zhong2025taming,fischer2025flowr,wei2025gsfix3d,yin2025gsfixer} propose some novel techniques to address the rendering artifacts of 3DGS, they predominantly employ diffusion models as post-processing tools for 2D rendering correction. Although they improves perceptual quality within observed regions, the generation results still lacks geometric awareness, resulting in catastrophic failures such as generating inconsistent depth, repeating textures, blurring artifacts when encountering large extrapolated areas.

Stemming from the fundamental ill-posedness of view extrapolation, we identify that the feedback loop between generation and reconstruction remains underexplored. Prior work treats diffusion refinement as a 2D post-process, overlooking a critical limitation that latent diffusion models lack 3D geometric priors knowledge. To bridge this gap, we introduce Leveling3D, the first framework that tightly couples feed-forward 3D reconstruction with geometry-conditioned diffusion generation. Built on AnySplat~\cite{jiang2025anysplat} and Stable Diffusion 2~\cite{Rombach_2022_SD2}, Leveling3D bridges reconstruction and generation through a geometry-aware adapter (Fig.~\ref{fig:overview}). Our key insights are: (1) transfering 3D geometric priors (tokens aggregated from multi-view inputs by VGGT~\cite{wang2025vggt}) to condition the diffusion model, grounding 2D generation in 3D structure; (2) the geometrically consistent extrapolated views generated by diffusion model are fed back to refine the 3DGS representation. This reconstruction-generation-refinement produces photorealistic details in underconstrained regions while preserving multi-view geometric consistency.
Our proposed pipeline comprises three key components: (i) a lightweight \emph{geometry-aware leveling adapter} that injects 3D geometric priors into diffusion via cross-attention; (ii) a \emph{palette filtering strategy} that ensure a well-behaved, approximately normal training distribution that prevents mode collapse (iii) a \emph{test-time masking refinement} that ensures seamless boundaries between reconstructed and generated regions. We achieve state-of-the-art performance on public unseen datasets across multiple scenes, including both the NVS rendering quality and geometry consistency, demonstrating that our geometry-aware approach significantly outperforms both vanilla diffusion model and other 3DGS fixing methods based on diffusion model.
Our contributions can be summarized as:
\begin{itemize}
    \item Leveling3D, a novel framework to tightly couple feed-forward 3D reconstruction with generation.
    \item Leveling adapter with a palette filtering training strategy and mask refinement that enables diffusion control with geometric priors for complex 3D-aware extrapolated view refinement.
    % \item A test-time masking refinement that produces artifact-free boundaries between reconstructed and generated regions.
    \item A refinement pipeline that leverages enhanced extrapolated views feedback to improve 3D representation quality.
    \item Extensive experiments demonstrate that Leveling3D outperforms existing SOTA on public datasets on both novel-view synthesis and depth estimation.
\end{itemize}

\section{Related Work}

\label{sec:related}
\noindent \textbf{Feed-Forward Novel-View Synthesis.} The shift from per-scene optimization to generalizable feed-forward networks has enabled 3DGS estimation without scene-specific training. Previous methods~\cite{charatan2024pixelsplat, chen2024mvsplat, xu2024depthsplat} explore pixel-aligned prediction by mapping 2D pixels to 3D Gaussians with transformers. While NoPoSplat~\cite{ye2024nopo}, Splatt3R~\cite{smart2024splatt3r}, and PF3plat~\cite{hong2024pf3plat} eliminated pose requirements, enabling zero-shot generalization from uncalibrated pairs. Recent models scale these ideas further, where FLARE~\cite{zhang2025flare} and AnySplat~\cite{jiang2025anysplat} unify sparse/dense uncalibrated collections, and VolSplat~\cite{wang2025volsplat} improved this with voxel-aligned prediction, directly regressing primitives from a 3D grid to avoid unreliable 2D matching. Despite these advances, all feed-forward methods fail in under-constrained regions where sparse observations yield geometric ambiguity.
\\
\\
% \paragraph{Sparse-View Novel-View Synthesis.} Sparse-view NVS requires hallucinating content from minimal observations. MVSplat~\cite{chen2024mvsplat} constructed plane-sweep cost volumes to establish cross-view correspondences, enabling reconstruction from two views. However, cost volumes struggle with extreme sparsity. MVSplat360~\cite{chen2024mvsplat360} addressed this by integrating feed-forward 3DGS with Stable Video Diffusion, rendering features into SVD's latent space for temporally consistent 360° synthesis from five views. PixelSplat~\cite{charatan2024pixelsplat} remains a calibrated baseline, while LEAP removes pose requirements via learned estimation. These methods reconstruct observed regions faithfully but cannot fill large missing areas while maintaining geometric consistency.
\noindent \textbf{Generation Control of Diffusion Model.} Controlling diffusion models beyond text prompts has become essential for spatial guidance. ControlNet~\cite{zhang2023ControlNet} explores trainable zero-convolutions for injecting structural conditions, while T2I-Adapter~\cite{mou2024t2i} provides a lighter-weight composable alternative. Dhariwal \etal~\cite{dhariwal2021diffusion} provides an additional noise-conditioned external classifier guidance for the denosing sampler. Training-free approaches like Universal Guidance~\cite{bansal2023universal} and FREEDOM~\cite{yu2023freedom} avoid per-condition training by steering generation via gradients from pre-trained models. However, all these methods operate purely in the 2D latent space and lack explicit geometric reasoning, ill-suited for maintaining multi-view consistency in 3D reconstruction. Our geometry-aware leveling adapter bridges this gap by directly conditioning on multi-view geometry prior, aligning diffusion priors with feed-forward 3D geometry to enable simultaneous reconstruction and generation.
\\
\\
\noindent \textbf{3D Reconstruction Enhancement with Generation.} Recent work has used diffusion models as post-processors to repair reconstruction artifacts. Difix3D+~\cite{wu2025difix3d+}  trained a single-step diffusion model on artifact-clean pairs and distilled priors into 3DGS optimization. GSFixer~\cite{yin2025gsfixer} advanced this with a DiT-based architecture that fuses 2D semantic and 3D geometric features from reference views to ensure multi-view consistency. GSFix3D~\cite{wei2025gsfix3d} similarly distilled diffusion priors using random mask augmentation. Moreover, Zhong \etal~\cite{zhong2025taming} used video diffusion to iteratively enhance views with the training-free guidance approach. However, these methods solely rely on image-based reference and operate as post-processing tools, failing at improving the underlying geometry representation. Leveling3D departs by tightly coupling reconstruction with geometry-aware diffusion in a unified framework.

\section{Method}
\subsection{Preliminary}

\begin{figure*}[t]
  \centering
  % \fbox{\rule{0pt}{3in} \rule{0.9\linewidth}{0pt}}
   \includegraphics[width=\linewidth]{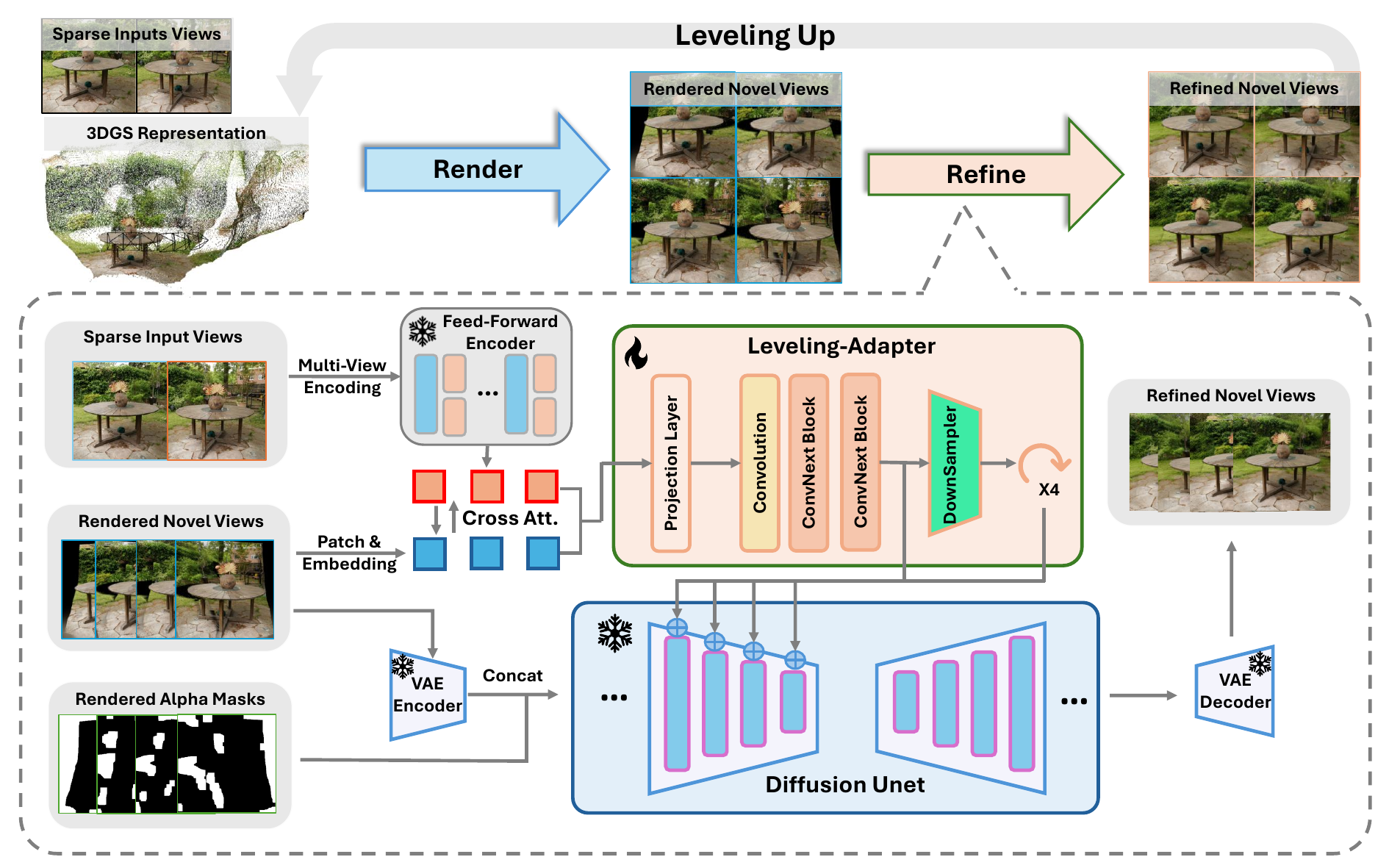}

   \caption{\textbf{Overview of the Leveling3D pipeline.} Our pipeline integrates a geometry-aware leveling adapter that fuses geometry tokens with diffusion control, enabling robust refinement of extrapolated views and leveling up the 3D reconstruction with extended geometric-consistent extrapolation areas.}
   \label{fig:overview}
\end{figure*}

\noindent\textbf{Feed-Forward 3D Gaussian Splatting.}
Feed-forward 3D reconstruction has emerged as a powerful alternative to per-scene optimization, enabling generalizable novel-view synthesis in a single forward pass. Here, we utilize AnySplat~\cite{jiang2025anysplat}, a feed-forward 3DGS model, as our baseline. Given $N$ view images $\{I_i\}_{i=1}^N$, where $I_i \in \mathbb{R}^{H\times W\times 3}$ of a 3D scene, Distilled from VGGT~\cite{wang2025vggt}, the geometry encoder from AnySplat~\cite{jiang2025anysplat} first aggregate tokens from the multi-view inputs. With the corresponding tokens, the camera head $F_C$ generates the poses $p_i$, the depth head $F_D$ produces the per-pixel depth map $D_i$ and a confidence map $C^D_i$. As in AnySplat~\cite{jiang2025anysplat}, the Gaussian primitives are extracted by the Gaussian head $F_G$ with the depth map $D_i$, poses $p_i$, and the aggregated tokens $\hat{t}^I$. To summarize, the feed-forward network reconstructs the scene as a set of 3D Gaussians and predicts the camera poses of input views as:
\begin{equation}
    f_{\boldsymbol{\theta}}:\{I_i\}_{i=1}^N
    \;\mapsto\;
    \Bigl\{
      \boldsymbol{\mu}, \sigma, \boldsymbol{r}, \boldsymbol{s}, \boldsymbol{c}
    \Bigr\}^\mathcal{G}
    \cup
    \{p_i\}_{i=1}^N.
    \label{eq:ffgs}
\end{equation}   Each Gaussian primitive $\mathcal{G}$ includes center $\boldsymbol{\mu}$, scaling factor $\boldsymbol{s}$, rotation $\boldsymbol{r}$, opacity $\sigma$, and color $\boldsymbol{c}$. As in~\cite{kerbl3Dgaussians}, given camera intrinsics and extrinsics, the 3DGS primitives are rasterized into per-pixel colors by:
\begin{equation}
\mathcal{C}(x_p) = \sum_{i\in K} \mathbf{c}i \alpha_i \prod_{j=1}^{i-1}(1 - \alpha_j), \; \alpha_i=\sigma_i\mathcal{G}^\prime(x_p)
\end{equation}
where $\mathcal{C}(x_p)$ is pixel color of the pixel position $x_p$,  $\mathcal{G}^\prime$ is the projected 2D Gaussian on image plane,  $K$ is the number of Gaussians correspond to the pixel.
\\
\\
\noindent\textbf{Stable Diffusion with Adapter Control.}
Stable Diffusion~\cite{Rombach_2022_SD2} is a latent diffusion model that performs denoising in a compressed latent space $z\in R$. The forward diffusion process progressively adds noise over $t\in T$ timesteps:
\begin{equation}
\mathbf{Z}_t = \sqrt{\bar{\alpha}_t} \mathbf{Z}_0 + \sqrt{1 - \bar{\alpha}_t} \boldsymbol{\epsilon}, \quad \boldsymbol{\epsilon} \sim \mathcal{N}(\mathbf{0}, \mathbf{I})
\end{equation}
where ${\alpha}_t$ defines the noise schedule. A UNet denoiser is learned by loss $L$ to reverse this process with condition $\mathbf{C}$:
\begin{equation}
\mathcal{L} = \mathbb{E}_{\mathbf{Z}_t,\mathbf{C}, t, \boldsymbol{\epsilon}} \left[ ||\boldsymbol{\epsilon} - \boldsymbol{\epsilon}_\theta(\mathbf{Z}_t, \mathbf{C})||_2^2 \right],
\end{equation}
where $\boldsymbol{\epsilon}_\theta$ is the Unet prediction. To inject control without retraining the full model, \cite{mou2024t2i} proposed a lightweight adapter $\mathcal{F}_{AD}$ to predict multi-scale condition features $\mathbf{F}_{c}=\{\mathbf{F}_c^1, \mathbf{F}_c^2, \mathbf{F}_c^3, \mathbf{F}_c^4\}$ that aligns with the intermediate features $\mathbf{F}_{enc}=\{\mathbf{F}_{enc}^1, \mathbf{F}_{enc}^2, \mathbf{F}_{enc}^3, \mathbf{F}_{enc}^4\}$ in the encoder of UNet denoiser. Given a condition $\mathbf{C}$ (e.g., coarse image, mask, and text prompt), the new features $\hat{\mathbf{F}}_{\text{enc}}$ are defined as the following:
\begin{equation}
\hat{\mathbf{F}}_{\text{enc}}^i = \mathbf{F}_{\text{enc}}^i + \mathbf{F}_c^i, \; \text{for } i \in {1,2,3,4}, \; \mathbf{F}_c = \mathcal{F}_{AD}(\mathbf{C})
\end{equation}

\subsection{Geometry-Aware Leveling Adapter}

\noindent\textbf{Adapter Control with Geometry Prior.} As shown in Fig.~\ref{fig:overview}, we design the novel Leveling Adapter as a lightweight network that bridges feed-forward 3D geometry with the diffusion model's latent space. Here, we define our goal of the Leveling Adapter as aligning the corresponding condition image with the geometry prior tokens from the multi-view geometry aggregation by AnySplat~\cite{jiang2025anysplat} encoder. Different from~\cite{mou2024t2i}, the Leveling Adapter includes one projection layer, four feature extraction blocks, and three downsample blocks. An $8\times8$ convolution with a stride of 8 is applied in the projection layer to align the condition from $512\times512$ to $64\times64$. In each feature extraction block, a convolution layer and two ConvNext blocks are used instead of residual blocks to extract spatial context while maintaining computational efficiency. After feature extraction, an average pooling is used for downsampling.
%, attending to the first token~\cite{barbero2025llms}, which is the most important
To capture the geometry information from the multi-view inputs, we utilize the output token of the geometry encoder. We first patchify the condition image $I_0$ into tokens $l_0=\frac{HW}{p^2}$ following VGGT~\cite{wang2025vggt}, where $p=14$ is the patch size, and $H, W$ are the height and width. Then, given the geometry token $\hat{t}_0^I$, we fuse the patchify tokens from the reference image with the geometry tokens by a cross-attention:

\begin{align}
\begin{split}
\left \{
\begin{array}{ll}
    \mathbf{Q} = \mathbf{W}_Q l_0;\ \mathbf{K} = \mathbf{W}_K l_0;\ \mathbf{V} = \mathbf{W}_V \hat{t}_0^I\\
    Attention(\mathbf{Q}, \mathbf{K}, \mathbf{V}) = softmax(\frac{\mathbf{Q}\mathbf{K}^T}{\sqrt{d}})\cdot \mathbf{V},
\end{array}
\right.
\end{split}
\end{align}
where $\mathbf{W}_Q$, $\mathbf{W}_K$, $\mathbf{W}_V\in \mathbb{R}^{768\times 768}$ are learnable projection matrices. Given the cross attention result, we unpatch it to generate the condition residual: \begin{equation}
\begin{split}
\mathbf{C}_{res} &= \text{Unpatch}\bigl( 
   \text{Attention}(\mathbf{Q}, \mathbf{K}, \mathbf{V})
\bigr)
\end{split}
\end{equation}The final condition is $\hat{\mathbf{C}}=\mathbf{C}+\mathbf{C}_{res}$, where $\mathbf{C}_{res}$ is the residual feature. Our adapter $\mathcal{F}_{level}$ extracts multi-scale features $\hat{\mathbf{F}}_c$ as:

\begin{equation}
\hat{\mathbf{F}}_{\text{enc}}^i = \mathbf{F}_{\text{enc}}^i + \hat{\mathbf{F}}_c^i, \; \text{for } i \in {1,2,3,4}, \; \hat{\mathbf{F}}_c = \mathcal{F}_{level}(\hat{\mathbf{C}})
\end{equation} Instead of naively incorporating the reference image as in Difix3D+~\cite{wu2025difix3d+}, our Leveling Adapter controls the diffusion process with the patchified tokens from geometry prior and reference. This design propagates information from confident, well-reconstructed geometry to the extrapolated areas with artifacts, yielding a geometry-aware condition that guides diffusion without corrupting the original reconstruction details.
\subsection{Training and Testing}

\noindent\textbf{Training Data Curation}. To establish our training dataset, we first select sparse sequences with a range of input numbers $N$ and the intervals $N_{in}$ between each input randomly. Then, the selected sequences are fed into the feed-forward reconstruction baseline~\cite{jiang2025anysplat} to generate the corresponding 3DGS representation. For each sequence, we rendered a set of novel-view images paired to the ground truth within the interval of $N_{in}$.
\\
\\
\noindent\textbf{Palette Filtering.}
As noted in~\cite{kim2025diffusehigh}, training generative models on statistically uninformative appearance distributions often leads to mode collapse, making it difficult to synthesize rich structural details. To mitigate this issue, we propose a \textit{palette filtering strategy} that samples training data whose masked-region intensity statistics are overly concentrated, so that the retained training set exhibits a more well-behaved (approximately normal, non-degenerate) distribution and provides more stable supervision. For each novel-view rendering, we obtain a transmittance map $O(x_p) = \sum_{i\in K} \alpha_i \prod_{j=1}^{i-1}(1 - \alpha_j)$. Following~\cite{zhong2025taming, zhong2025empowering} and derive an opacity mask $\bar{M} = \{ O < \eta_{\text{mask}} \}$ by thresholding the transmittance map with $\eta_{\text{mask}}$. We then analyze the intensity distribution inside $\bar{M}$ using the intensity map $I_{\text{int}}$.
Let $\mu$ and $\sigma$ denote the mean and standard deviation of $\{I_{\text{int}}(\mathbf{p}) \mid \mathbf{p}\in\bar{M}\}$.
We define the \textit{palette score} $\mathcal{S}_{\text{p}}$ as the fraction of masked pixels whose intensities fall within one standard deviation of the mean:
\begin{equation}
\mathcal{S}_{\text{p}} = \frac{1}{|\bar{M}|} \sum_{\mathbf{p} \in \bar{M}} \mathbf{1} \left[ \big| I_{\text{int}}(\mathbf{p}) - \mu \big| < \sigma \right],
\label{eq:palette_score}
\end{equation}
where $\mathbf{1}[\cdot]$ is the indicator function. By definition of $1\sigma$ rule, approximately 68\% of the probability mass in a normal distribution lies within one standard deviation of the mean. Motivated by this property, we set $\eta_{\text{p}} = 0.68$ as an optimal baseline to sample the training data with the informative regions $\mathcal{S}_{\text{p}} > \eta_{\text{p}}$, preventing an overly simple appearance that provides limited learning signal.
\\
\\
\noindent\textbf{Optimization}. To train the leveling adapter, we accumulate the gradients of each training pair from a sequence that shares the same geometry tokens for backpropagation, ensuring geometry consistency. Finally, we formulate the training loss for training a paried sequence of $N$ as:
\begin{equation}
    L_N = \frac{1}{N}\sum_{i\in N} L_{\text{LPIPS}}({I}_i, \tilde{I}_i),
\end{equation} where $L_{\text{LPIPS}}$ is the \textit{LPIPS} perceptual loss with VGG-16~\cite{simonyan2014very}, $\tilde{I}_i$ is the refined diffusion model output from $\bar{I}_i$, and ${I}_i$ is the paired ground truth.
\\
\\
\noindent\textbf{Test-Time Mask Refinement.} To achieve more satisfactory refinement details, we adopt the stable diffusion model~\cite{Rombach_2022_SD2} finetuned on the inpainting task as our generation baseline, where the opacity mask is applied for the inpainting area. However, directly incorporating the mask $M$ leads to messy boundaries between reconstructed and generated regions. Therefore, we employ a test-time masking refinement strategy to resolve this challenge. With the original opacity mask $\bar{M}$, we define the test-time mask refinement process as a combination of morphological operations:
    
\begin{equation}
    \begin{aligned}
    {M}^{\text{close}} &= 1-((\bar{M} \oplus \mathbf{K}_{5}) \ominus \mathbf{K}_{5}), \\
    {M}^{\text{refine}} &= {M}^{\text{close}} \oplus \mathbf{K}_{20},
    \end{aligned}
    \label{eq:mask}
\end{equation}
 where $\oplus$ and $\ominus$ denote morphological dilation and erosion operator, $\mathbf{K}_{n}$ is an $n\times n$ structuring kernel. The closed mask, $M^{\text{close}}$, is obtained by applying a morphological closing operation followed by a hole-filling procedure. To expand the mask boundaries, we further apply dilation to generate a refined mask, $M^{\text{refine}}$. We evaluate various combinations of kernel sizes on the ScanNet dataset to determine the optimal configuration; detailed results are provided in Table \ref{tab:ablation_kernel}. This refinement during testing yields a soft mask that identifies under-constrained regions, ensuring spatial continuity and mitigating boundary artifacts.

\begin{figure}[t]
% \vspace{-1em}
\begin{algorithm}[H]
\caption{Feed-Forward 3DGS Refinement with Generation.}
\label{alg:optim}
\begin{algorithmic}[1]
\STATE \textbf{Function:} Feed-Forward Reconstruction Baseline $f_{\theta}(\cdot)$, generation model $\mathbf{G}(\cdot)$.
\STATE \textbf{Input: }Sparse inputs of N images $\{I_i\}_{i=1}^N$. $L$ Novel-view poses $\bar{p}_i, i\in L$
\STATE \textbf{Output: }Refined, completed 3DGS $\hat{\mathcal{G}}$. 
\STATE \textbf{Variable:} List of high-quality images $\mathbf{H}=[\;]$
\STATE Feed-forward reconstruction to generate 3DGS $f_\theta(I_{i\in N})\Rightarrow \mathcal{G}$ \hfill $\triangleright$ Eq.~\eqref{eq:ffgs}
\STATE Aggregate the geometry tokens $\hat{t}^I$.
\FOR{$i = 0, \ldots, N-1$}
    \STATE Append ${I}_i$ to $\mathbf{H}$
\ENDFOR
\FOR{$i = 0, \ldots, L-1$}
    \STATE Render image and mask from 3DGS: $\bar{I}_i, \bar{M}_i = \text{rasterize}(\mathcal{G}, \bar{p}_i)$
    \STATE Test-time mask refinement $\bar{M}_i \Rightarrow{{M}}^{\text{refine}}_i$ \hfill $\triangleright$ Eq.~\eqref{eq:mask}
    \STATE Refine extrapolation views with generation: $\tilde{I}_i = \mathbf{G}(\bar{I}_i, M^{\text{refine}}_i, I, \hat{t}^I)$
    \STATE Append $\tilde{I}_i$ to $\mathbf{H}$
\ENDFOR
\STATE Leveling up the feed-forward reconstruction with refined inputs $f_\theta(\mathbf{H})\Rightarrow \hat{\mathcal{G}}$ 
\end{algorithmic}
\end{algorithm}
\end{figure}
\subsection{Refine 3D Reconstruction with Generation}
The key insight of Leveling3D is that generation-enhanced views can serve as the refined inputs to improve the feed-forward 3DGS reconstruction by a leveling-up process. Prior works~\cite{wu2025difix3d+, wei2025gsfix3d, yin2025gsfixer, viewextrapolator} generally suffer from two critical limitations: (i) rendering fixing fails for novel extrapolated views far from input views, where geometric priors become unreliable, and (ii) requires an iterative update with loss backpropagation that is computationally expensive and slow. Our approach addresses both by performing geometry-aware enhancement pipeline, achieving broader applicability and significantly higher efficiency.

With the set of sparse input images $\{I\}$, we generate the coarse 3DGS representation $\mathcal{G}$ from the feed-forward baseline. Following~\cite{zhong2025taming}, we subsample a set of extrapolated poses $\{\bar{p}\}$ between and near the sparse inputs to render a set of coarse extrapolated NVS results $\{\bar{I}\}$. After generating a refined set of extrapolated views $\{\tilde{I}\}$, we update the 3DGS representation $\hat{\mathcal{G}}$ by feeding $\{I\} \cup \{\tilde{I}\}$ into the feed-forward 3DGS. This feedback then refines both geometry and appearance, resulting in a more complete 3DGS representation. We provide the pipeline as pseudo code in Alg.~\ref{alg:optim}, which illustrates the update step in Leveling3D.

\section{Experiments}
\subsection{Settings}
\label{sec:eval-in-the-wild}
% \paragraph{Datasets and Metrics.} We train on a random selection of 80\% of scenes (112 out of a total of 140) from the DL3DV~\cite{ling2024dl3dv} benchmark dataset. We generate 80,000 noisy-clean image pairs using the dataset curation strategies listed in \cref{tab:data_curation}, and simulate NeRF and 3DGS-based artifacts in a 1:1 ratio.

% \paragraph{Baselines.} We evaluate  with Nerfacto~\cite{tancik2023nerfstudio} and 3DGS~\cite{kerbl3Dgaussians} backbones on the 28 held out scenes from the DL3DV~\cite{ling2024dl3dv} benchmark and the 12 captures in the Nerfbusters~\cite{warburg2023nerfbusters} dataset. We partition each scene into a set of reference views used during training and evaluate on the left-out target views. We generate these splits for DL3DV by partitioning frames into two clusters based on camera position, ensuring a substantial deviation between reference and target views. We select reference and target views in the Nerfbusters dataset following their recommended protocol~\cite{warburg2023nerfbusters}.
\noindent\textbf{Datasets and Metrics.}
We generate training data from DL3DV~\cite{ling2024dl3dv} (10,510 scenes) and ScanNet++~\cite{yeshwanth2023scannet++} (1,006 scenes), randomly selecting 80\% of scenes to produce $\sim$100,000 noisy-clean image pairs. All baselines are tested on unseen datasets: MipNeRF360~\cite{barron2022mip} and VRNeRF~\cite{xu2023vr} for novel-view synthesis, TartanAir~\cite{wang2020tartanair} (synthetic indoor/outdoor), and ScanNet~\cite{dai2017scannet} (real-world indoor) for novel-view depth estimation.  We conduct all experiments using the common metrics such as PSNR, SSIM, and LPIPS for NVS. In terms of depth estimation, we use Abs Rel, RMSE, $\delta<1.25$ following~\cite{wang2025pi3}, and further apply Met3R~\cite{asim25met3r} for generative multi-view 3D consistency evaluation. Results are averaged over the extrapolated novel views rendered from the feed-forward 3DGS representation after refinement. for each dataset. All experiments are conducted in the sparse two-view input setting, except for different numbers of inputs. More details are available in the supplementary.
\\
\\
\noindent\textbf{Baselines.} We evaluate our method against state-of-the-art diffusion-based enhancement pipelines, categorized into image-diffusion and video-diffusion approaches. For the former, we compare against {Difix3D+}~\cite{wu2025difix3d+} and {GSFix3D}~\cite{wei2025gsfix3d}; for the latter, we adopt {GSFixer}~\cite{yin2025gsfixer} and {ViewExtrapolator}~\cite{viewextrapolator} as our primary baselines. All methods are evaluated on both novel-view synthesis and novel-view depth estimation tasks. To ensure a fair comparison, all diffusion-based baselines utilize AnySplat as the underlying pose-free 3DGS feed-forward framework.
% For both training and testing, we sample image sequences with start/end frames as inputs and 4--$N$ intermediate frames (with ground-truth poses) from a single scene. Inputs are processed by Anysplat~\cite{} to initialize Gaussian primitives and render initial novel views. Besides, we will use the first input image and its corresponding VGGT transformer tokens as the reference of this sequence.
% During training, the rendered (noisy) and ground-truth (clean) intermediate images form paired data to supervise the Geometry-Aware Leveling Adapter with the input of reference data, which learns to refine Anysplat’s initial outputs.
% At inference, we apply iterative refinement: Anysplat generates initial views $\to$ the trained adapter refines them $\to$ refined views are fed back with original inputs into Anysplat for final optimization, producing high-quality novel views and depth.
\\
\\
\noindent\textbf{Implementation Details.} We train the model on 8× NVIDIA A6000 GPUs with 2 batch size for each using the AdamW optimizer (learning rate: 1e-4, weight decay: 1e-2) for 20 epochs. During training, we freeze both AnySplat and Diffusion Model, optimizing only the parameters of the Leveling Adapter. Additional details for extrapolated poses sampling and experiments for the palette filtering are in the supplementary.

\begin{table*}[t]
  \centering
  \caption{\textbf{Quantitative comparison on MipNeRF360~\cite{barron2022mip} and VRNeRF~\cite{xu2023vr} datasets with sparse two-views input}. The best result is in \textbf{bold}, and the second-best is \underline{underlined}. All experiments are conducted with the same baseline~\cite{jiang2025anysplat}.}
  \resizebox{0.9\linewidth}{!}{
  \begin{tabular}{@{}l|c|ccc|ccc|c@{}}
    \toprule
    \multirow{2}{*}{Methods}& \multirow{2}{*}{\makecell[l]{Diffusion\\Type}}& \multicolumn{3}{c|}{MipNeRF360~\cite{barron2022mip} Dataset} & \multicolumn{3}{c|}{VRNeRF~\cite{xu2023vr} Dataset} &  \multirow{2}{*}{Time(s)} \\
    & & PSNR$\uparrow$ & SSIM$\uparrow$ & LPIPS$\downarrow$  & PSNR$\uparrow$ & SSIM$\uparrow$ & LPIPS$\downarrow$  \\
    \midrule
    Baseline~\cite{jiang2025anysplat} & N/A& 15.60 & 0.318 & 0.314  & 15.89 & 0.532 & \underline{0.347} & N/A  \\
    \midrule
    
    ViewExtrapolator~\cite{viewextrapolator}&\multirow{2}{*}{Video} & 14.85 & 0.324 & 0.606 & \underline{16.716} & \underline{0.591} & 0.518 & 2.59 \\
    
    GSFixer~\cite{yin2025gsfixer} && \underline{15.69} & 0.332 & 0.348 & 15.86 & 0.554 & 0.356 & 12.07 \\
    \midrule
    GSFix3D~\cite{wei2025gsfix3d} && 14.96 & 0.302 & 0.354 & 13.47 & 0.412 & 0.53 & \underline{1.094} \\
    Difix3D+~\cite{wu2025difix3d+}&\multirow{1}{*}{Image} & 15.68 & \underline{0.334} & \underline{0.312}  & 16.22 & 0.540 & 0.363 & \textbf{0.718} \\
    Ours && \textbf{16.76} & \textbf{0.352} & \textbf{0.306}  & \textbf{18.35} & \textbf{0.610} & \textbf{0.316} & 1.167  \\
    \bottomrule
  \end{tabular}
  }
  % \vspace{2mm}
  % \vspace{-3mm}
  \label{tab:main_results}
\end{table*}
\begin{figure*}[t!]
  \centering
  % \fbox{\rule{0pt}{4in} \rule{0.9\linewidth}{0pt}}
   \includegraphics[width=\linewidth]{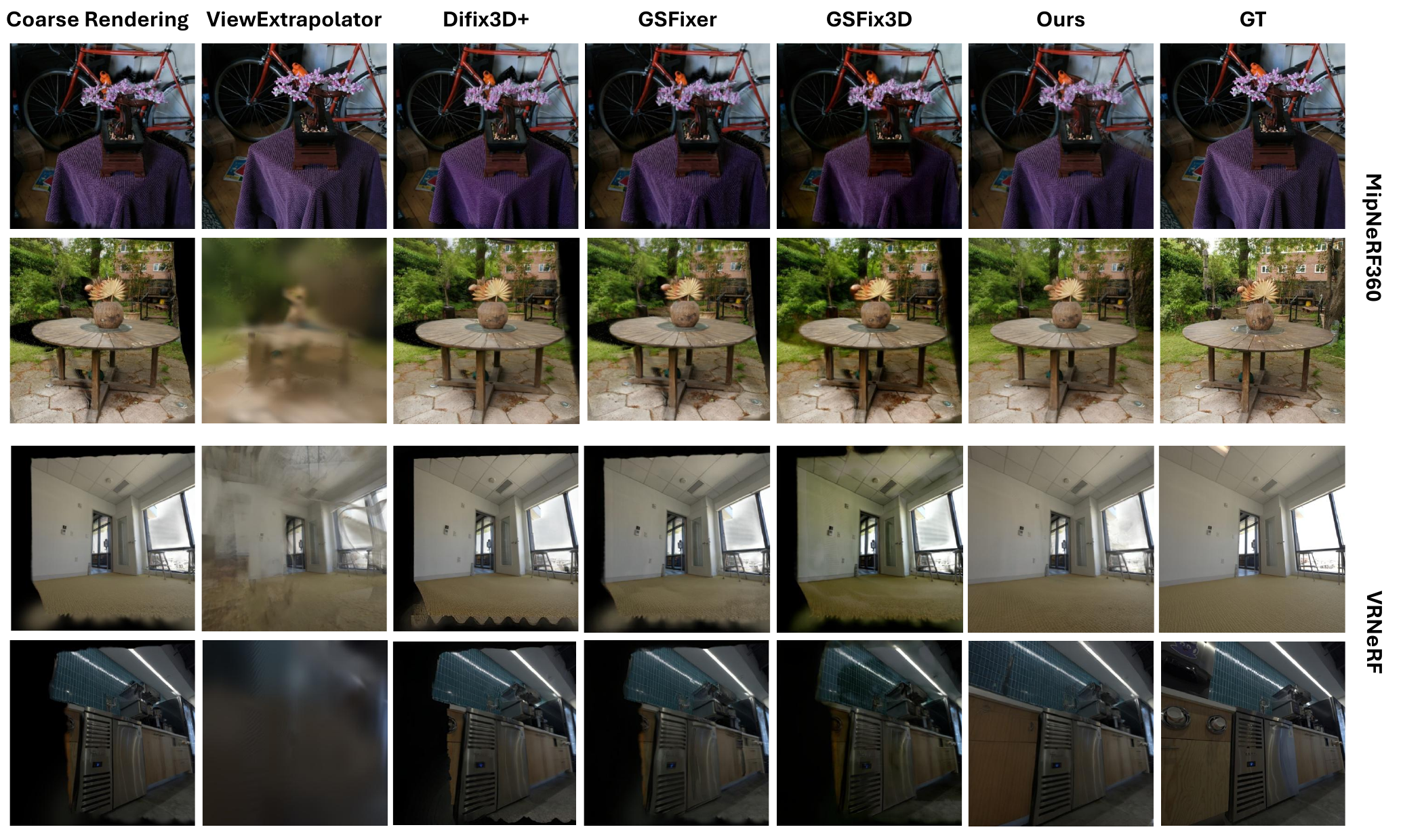}
   \caption{\textbf{Qualitative result on MipNeRF360~\cite{barron2022mip} and VRNeRF~\cite{xu2023vr} datasets.} Previous refinement methods suffer from severe geometry collapse, texture corruption, and missing content. In contrast, our Leveling3D method robustly generate fine details and fills large extrapolated areas with plausible content.}
   \label{fig:NVS_mip_VR}
\end{figure*}

\subsection{Main Results}
\noindent\textbf{Novel-View Image Synthesis.} We evaluate Leveling3D on real-world sparse captures from the subset of the unseen MipNeRF360~\cite{barron2022mip} and VRNeRF~\cite{xu2023vr} dataset. As shown in Table~\ref{tab:main_results}, our method achieves all SOTA results of all metrics, outperforming AnySplat~\cite{jiang2025anysplat} by 7.5\% on MipNeRF360 / 15.5\% on VRNeRF, Image-Diffusion based method (Difix3D+~\cite{wu2025difix3d+} and GSFix3D~\cite{wei2025gsfix3d}) by at least 6.8\% on MipNeRF360 / 13.1\% on VRNeRF and Video-Diffusion based method (GSFixer~\cite{yin2025gsfixer} and ViewExtrapolator~\cite{viewextrapolator}) by at least 6.7\% on MipNeRF350 / 9.7\% on VRNeRF. Qualitative results in Fig.~\ref{fig:NVS_mip_VR} show robust recovery of texture-rich surfaces for challenging extrapolation views. Here, we conduct an additional step update for all above methods. While achieving high accuracy, Leveling3D maintains an inference speed comparable to image-based diffusion methods and significantly faster than video-based diffusion counterparts, demonstrating a favorable accuracy-efficiency trade-off.
\\
\\

\begin{table*}[t]
  \centering
  \caption{ \textbf{Quantitative results of depth estimation on TartanAir~\cite{wang2020tartanair} and ScanNet~\cite{dai2017scannet} datasets}. We evaluate all methods with the same baseline~\cite{jiang2025anysplat}, the same sparse two-view input, and novel poses. Our method outperforms other baselines.}
  \resizebox{\linewidth}{!}{
  \begin{tabular}{@{}l|cccc|cccc@{}}
    \toprule
    \multirow{2}{*}{Methods}& \multicolumn{4}{c|}{TartanAir~\cite{wang2020tartanair} Dataset} & \multicolumn{4}{c}{ScanNet~\cite{dai2017scannet} Dataset} \\
    & Abs Rel $\downarrow$ & RMSE $\downarrow$ & $\delta<1.25$ $\uparrow$ & Met3R $\downarrow$ & Abs Rel $\downarrow$ & RMSE $\downarrow$ & $\delta<1.25$ $\uparrow$ & Met3R $\downarrow$ \\
    \midrule
    Baseline~\cite{jiang2025anysplat} & 0.915 & 20.748  & 0.311 & 0.0683 & 0.372 & 58.826 & 0.708 & 0.0472 \\

    ViewExtrapolator~\cite{viewextrapolator} & 3.791 & 139.387  & 0.102 & 0.0633 & 3.794 & 120.666 & 0.323 & 0.0646 \\

    GSFix3D~\cite{wei2025gsfix3d} & 0.905 & 20.493  & 0.315 & 0.0675 & 0.29 & 33.398  & \underline{0.791} & 0.0453\\

    GSFixer~\cite{yin2025gsfixer} & \underline{0.892} & 20.366 & 0.315 &  0.0665 & 0.296 & 31.296 & 0.784 & 0.0462  \\
    Difix3D+~\cite{wu2025difix3d+} & 0.893 & \underline{20.298} &  \underline{0.321} & \underline{0.0625} & \underline{0.286} &  \underline{0.358} & 0.787 & \underline{0.0452} \\
    
    Ours & \textbf{0.853} & \textbf{19.54} & \textbf{0.351} & \textbf{0.0614} & \textbf{0.252} & \textbf{24.68}  & \textbf{0.826} & \textbf{0.0376} \\
    \bottomrule
  \end{tabular}
  }
  \vspace{2mm}
  
  % \vspace{-3mm}
  \label{tab:depth_results}
\end{table*}
\begin{figure*}[t!]
  \centering
  % \fbox{\rule{0pt}{1.5in} \rule{0.9\linewidth}{0pt}}
   \includegraphics[width=\linewidth]{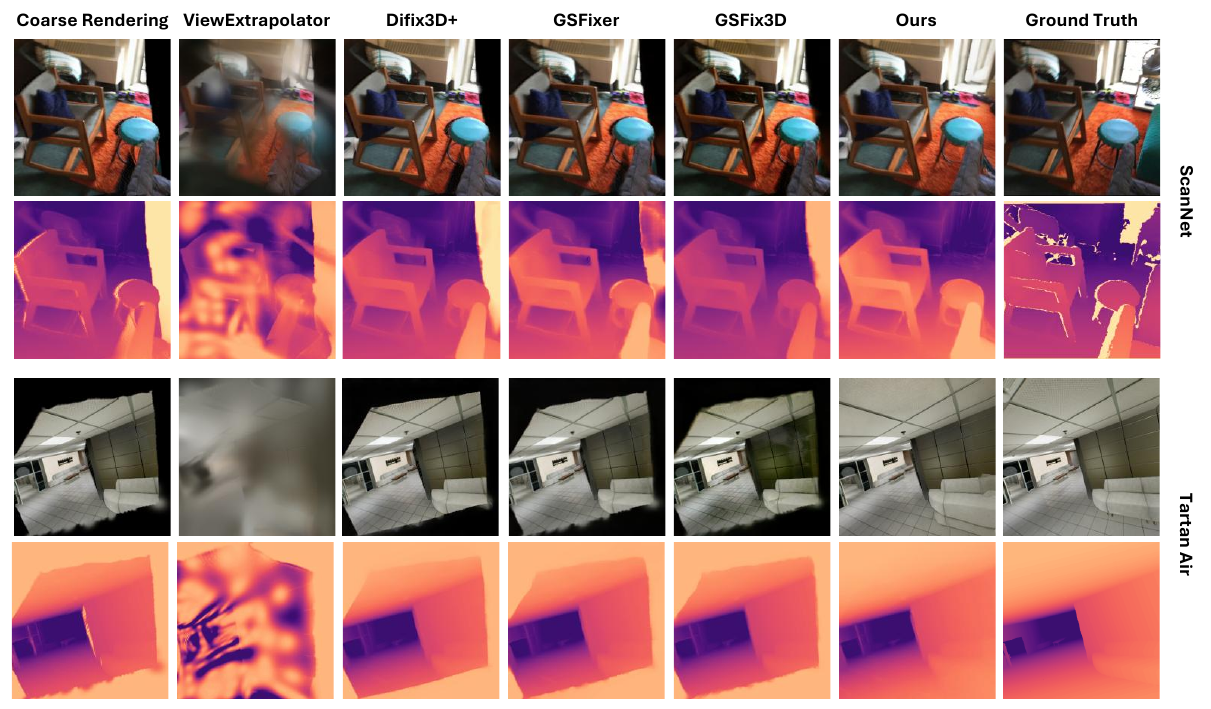}
   \caption{\textbf{Qualitative result of depth estimation on TartanAir~\cite{wang2020tartanair} and ScanNet~\cite{dai2017scannet} datasets.} Our method refines geometry-consistent novel views within the extrapolated artifact areas, leveling up the 3DGS representation to achieve complete scene reconstruction with superior geometry detail recovery and boundary coherence.}
   \label{fig:depth_result}
\end{figure*}
\noindent\textbf{Novel-View Depth Estimation.} We also evaluate the novel-view depth estimation with sparse two-view inputs setting. As shown in Table~\ref{tab:depth_results}, our method reduces AbsRel by 9\% on TartanAir / 34\% on ScanNet compared with AnySplat~\cite{jiang2025anysplat}, at least 5.3\% on TartanAir / 12.4\% compared with image-diffusion based methods (Difix3D+~\cite{wu2025difix3d+} and GSFix3D~\cite{wei2025gsfix3d}) and at least 4.9\% on TartanAir / 17.4\% on ScanNet compared with video-diffusion based methods (GSFixer~\cite{yin2025gsfixer} and ViewExtrapolator~\cite{viewextrapolator}). Furthermore, our method achieves the lowest Met3R~\cite{asim25met3r} scores on both datasets (0.0614 on TartanAir and 0.0376 on ScanNet), indicating that our method achieves a superior multi-view geometric consistency across generated views. Fig.~\ref{fig:depth_result} shows sharper object boundaries and plausible completion in the artifact regions of the extrapolated view instead of over-smoothed or erroneous depth from Difix3D+. The results indicate that the Leveling Adapter refines novel views to provide denser geometric cues for a more complete feed-forward 3D reconstruction.

\begin{table*}[t]
\centering
\caption{\textbf{Quantitative evaluation of view-dependency} on the ScanNet dataset. 
We compare our method against AnySplat across varying numbers of input views. 
Bold numbers indicate the best performance.}
\label{tab:num_input_views_results}

\resizebox{0.7\linewidth}{!}{
\begin{tabular}{l|ccc|ccc}
  \toprule
  & \multicolumn{3}{c|}{\textbf{Baseline~\cite{jiang2025anysplat}}} 
  & \multicolumn{3}{c}{\textbf{Ours}} \\
  \textbf{Views} & PSNR$\uparrow$ & SSIM$\uparrow$ & LPIPS$\downarrow$ 
  & PSNR$\uparrow$ & SSIM$\uparrow$ & LPIPS$\downarrow$ \\
  \midrule
  2-Views  & 9.506  & 0.176 & 0.804 & \textbf{12.296} & \textbf{0.392} & \textbf{0.776} \\
  3-Views  & 10.076 & 0.272 & 0.722 & \textbf{12.676} & \textbf{0.404} & \textbf{0.702} \\
  4-Views  & 10.794 & 0.288 & 0.724 & \textbf{13.272} & \textbf{0.412} & \textbf{0.704} \\
  5-Views  & 11.242 & 0.322 & 0.708 & \textbf{13.446} & \textbf{0.424} & \textbf{0.678} \\
  6-Views  & 11.638 & 0.344 & 0.686 & \textbf{14.041} & \textbf{0.442} & \textbf{0.664} \\
  7-Views  & 12.266 & 0.376 & 0.642 & \textbf{14.478} & \textbf{0.481} & \textbf{0.624} \\
  8-Views  & 12.592 & 0.404 & 0.632 & \textbf{14.592} & \textbf{0.484} & \textbf{0.611}  \\
  10-Views & 12.951 & 0.418 & 0.623 & \textbf{14.606} & \textbf{0.491} & \textbf{0.606} \\
  \bottomrule
\end{tabular}
}

\label{tab:different_num_view}
\end{table*}

\subsection{Comparison on Different Number of Sparsities.} To evaluate the robustness of Leveling3D to varying sparsity levels, we conduct experiments on the ScanNet~\cite{dai2017scannet} dataset with 2-10 sparse input views per scene. Table~\ref{tab:num_input_views_results} summarizes the results. For each sparsity level, we evaluate all metrics averaged on all extrapolated views rendered from the refined feed-forward 3DGS. Our method demonstrates improvement across all sparsities compared with the baseline. With 2 views (extremely sparse), Leveling3D achieves 12.296 PSNR, a 22.6\% improvement over 9.50. As inputs increase to 10 views, gaps narrow, but ours still leads by 14.606 PSNR, which is 12\% over 12.951 PSNR of the baseline, highlighting the adapter's lightweight efficiency and superior generalization ability across different sparsities.

\begin{table*}[t]
\centering

% 左侧：Kernel size ablation
\begin{minipage}[t]{0.49\textwidth}
  \centering
  \caption{\textbf{Ablation on the mask refinement kernel size.} 
    The best result is highlighted in \textbf{bold}, and the second-best is \underline{underlined}. We choose (5, 20) as our optimal configuration, which achieves the best PSNR / SSIM / LPIPS.}
  \label{tab:ablation_kernel}
  % A smaller kernel fails to fully encompass messy boundaries, while a larger kernel erroneously encroaches upon well-reconstructed regions, causing unnecessary modifications to ground-truth pixels that should remain untouched. 
  \resizebox{0.98\linewidth}{!}{
    \begin{tabular}{@{}l|cccc@{}}
      \toprule
      (Clos., Dila.) & PSNR $\uparrow$ & SSIM $\uparrow$ & LPIPS $\downarrow$ \\
      \midrule
      (5,10)  & 16.756 & 0.558 & \underline{0.417} \\
      (5,20)  & \textbf{17.017} & \textbf{0.562} & \textbf{0.417} \\
      (5,30)  & \underline{17.011} & \underline{0.561} & 0.422 \\
      (10,10) & 16.637 & 0.555 & 0.420 \\
      (10,20) & 16.897 & 0.559 & 0.419 \\
      (10,30) & 16.918 & 0.558 & 0.423 \\
      (20,10) & 16.509 & 0.554 & 0.422 \\
      (20,20) & 16.779 & 0.557 & 0.420 \\
      (20,30) & 16.820 & 0.556 & 0.424 \\
      \bottomrule
    \end{tabular}
  }
\end{minipage}%
\hfill
% 右侧整体：上下两个表
\begin{minipage}[t]{0.49\textwidth}
  \centering
  
  % 右上：不同控制方法的比较
  \begin{minipage}[t]{\linewidth}
    \centering
    \caption{\textbf{Comparison of different generative control.} *: naive diffusion model.}
    \label{tab:different_diffusion}
    %      * represents baseline refined with diffusion model without control.
    \resizebox{0.95\linewidth}{!}{
      \begin{tabular}{@{}l|cccc@{}}
        \toprule
        Method & PSNR$\uparrow$ & SSIM$\uparrow$ & LPIPS$\downarrow$ & FID$\downarrow$ \\
        \midrule
        Baseline~\cite{jiang2025anysplat} & 13.21 & 0.464 & 0.472 & \underline{177.91} \\
        w/o control* & 15.13 & 0.506 & 0.494 & 256.15 \\
        T2I~\cite{mou2024t2i} & \underline{16.58} & \underline{0.554} & \underline{0.446} & 186.73 \\
        ControlNet~\cite{zhang2023ControlNet} & 15.51 & 0.542 & 0.464 & 197.00 \\
        Ours & \textbf{16.89} & \textbf{0.562} & \textbf{0.414} & \textbf{156.62} \\
        \bottomrule
      \end{tabular}%
    }
  \end{minipage}
  
  \vspace{2mm}   % 上下表之间的垂直间距，可根据实际效果微调 (0.8ex ~ 1.5ex 较自然)
  
  % 右下：VRNeRF 逐模块消融
  \begin{minipage}[t]{\linewidth}
    \centering
    \caption{\textbf{Ablation study}. (a) Naive diffusion, (b) Geometry token fusion, (c) Palette filtering, (d) Mask refinement.}
      %  on VRNeRF~\cite{xu2023vr} dataset
    \label{tab:ablation_stages}
    
    \resizebox{0.98\linewidth}{!}{
      \begin{tabular}{@{}l|ccccc@{}}
        \toprule
        Method & PSNR$\uparrow$ & SSIM$\uparrow$ & LPIPS$\downarrow$ & FID$\downarrow$ & Met3R$\downarrow$\\
        \midrule
        Baseline~\cite{jiang2025anysplat} & 15.89 & \underline{0.532} & 0.354 & 153.39 & 0.0322 \\
        +(a)      & 18.19 & 0.372 & 0.336 & 169.78 & 0.0264 \\
        +(a)+(b)  & \underline{18.35} & 0.380 & 0.320 & 146.80 & \underline{0.0248} \\
        +(a)+(b)+(c) & 18.27 & 0.392 & \underline{0.316} & \underline{141.41} & 0.0252 \\
        +(a)+(b)+(c)+(d) & \textbf{18.36} & \textbf{0.610} & \textbf{0.314} & \textbf{138.31} & \textbf{0.0242} \\
        \bottomrule
      \end{tabular}%
    }
  \end{minipage}
  
\end{minipage}

\end{table*}

\begin{figure}[t!]
  \centering
  % \fbox{\rule{0pt}{1.5in} \rule{0.9\linewidth}{0pt}}
   \includegraphics[width=\linewidth]{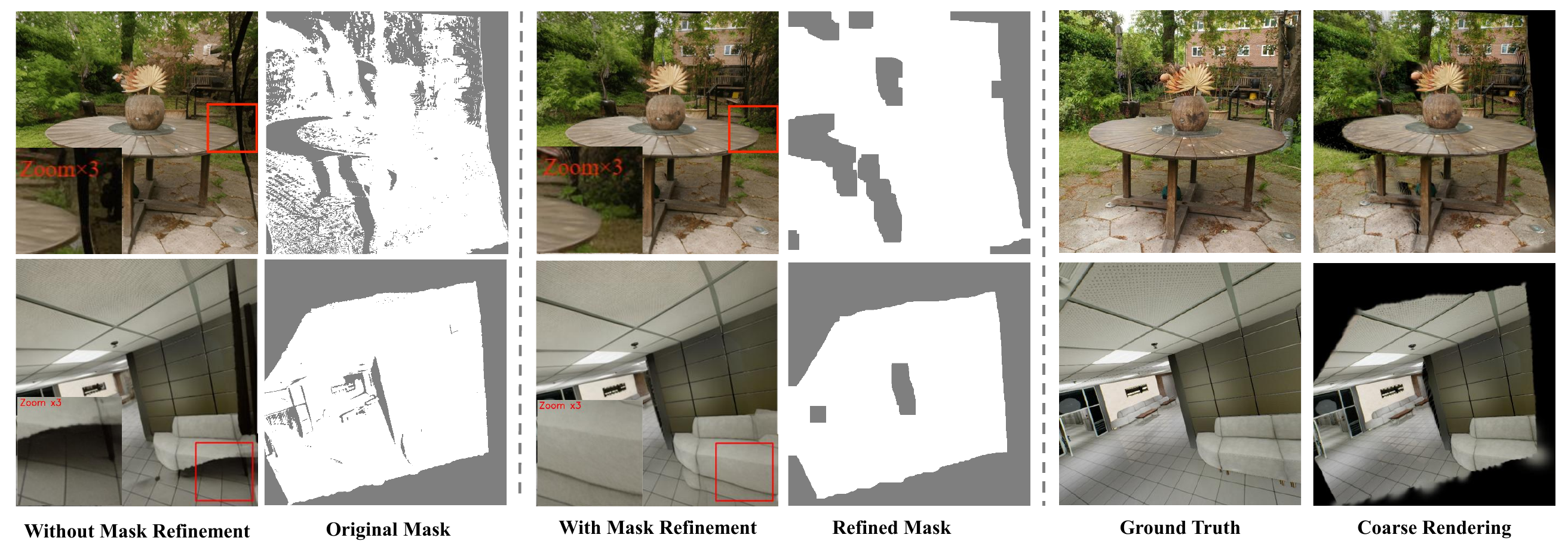}
   \caption{\textbf{Qualitative result by removing the test-time mask refinement.} Our mask refinement robustly prevent corrupted generation near the boundary.}
   \label{fig:mask_refine}
\end{figure}

\begin{figure}[t]
  \centering
  % \fbox{\rule{0pt}{1.5in} \rule{0.9\linewidth}{0pt}}
   \includegraphics[width=0.9\linewidth]{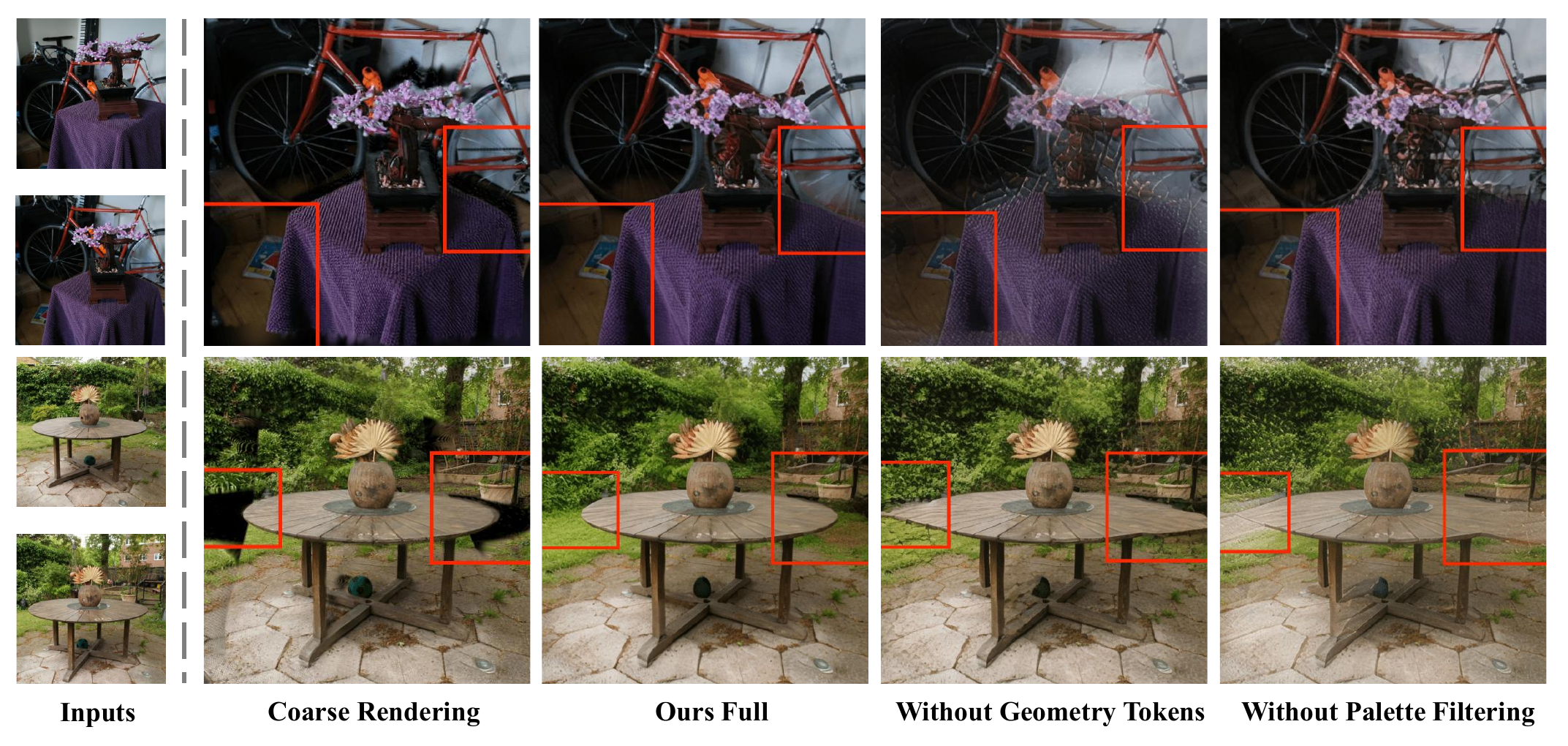}

   \caption{\textbf{Qualitative result by removing geometry tokens and palette filtering.} Our full pipeline generate sharp details and coherent geometry.}
   \label{fig:token_palette}
\end{figure}

\subsection{Ablation Study}
\noindent\textbf{Mask Refinement Kernel Size.} To investigate the impact of the mask refinement module, we conduct an ablation study on ScanNet~\cite{dai2017scannet} for the kernel sizes of closing and dilation operations ($\mathbf{K}_{close}$ and $\mathbf{K}_{dilat}$). As reported in Table ~\ref{tab:ablation_kernel}, the refinement performance is highly sensitive to the spatial extent of the mask. A small kernel ($\mathbf{K}_5$) is sufficient for the closing operation to fill internal holes within the artifact areas. However, for the dilation process, a moderate kernel size ($\mathbf{K}_{20}$) is essential to ensure that the diffusion model's guidance covers the entire artifacts at the boundaries. Crucially, we observe that an overly aggressive dilation  (e.g., $n > 30$) leads to a decline in reconstruction quality. This is because a large kernel causes the mask to encroach upon the high-fidelity regions originally rendered by the feed-forward 3DGS, resulting in the unnecessary replacement of accurate geometric structures with generated content. We evaluated this across diverse testing scenes under various conditions, identifying the optimal combination for the 448 × 448 resolution. Our optimal configuration strikes a balance between eliminating artifacts and preserving the integrity of the initial 3D reconstruction.
\\
\\
\noindent\textbf{Different Generation Control Methods.} We compare Leveling3D against different generation control methods on the ScanNet~\cite{dai2017scannet} dataset. Quantitative results are shown in Table~\ref{tab:different_diffusion}. On the ScanNet dataset, our method achieves 16.89 PSNR, outperforming ControlNet~\cite{zhang2023ControlNet}'s PSNR 15.51 and T2I~\cite{mou2024t2i} Adapter's PSNR 16.58. In terms of cross-view generation ability, our method achieves the lowest FID. These results demonstrate the superiority of geometry prior with sparse inputs, highlighting the robustness for extrapolated-view refinement.
\\
\\
\noindent\textbf{Proposed Components.} We ablate each proposed key component on VRNeRF~\cite{xu2023vr}: (a) geometry-aware leveling adapter, (b) geometry token fusion, (c) palette filtering, and (d) test-time mask refinement. As shown in Table~\ref{tab:ablation_stages}, without this mask refinement, the SSIM experiences a severe drop compared to the baseline, because the unconstrained diffusion process overwrites the high-fidelity structural details as shown in \cref{fig:mask_refine}. Without palette filtering, model collapse reduces SSIM by 3\%, failing on complex patterns. Disabling geometry token fusion lowers SSIM by 2\%. Finally, omitting the leveling adapter decreases PSNR by 14\%.
Qualitative results for geometry tokens and palette filtering are in \cref{fig:token_palette}.

\section{Conclusion}
In this work, we present Leveling3D, a novel framework that bridges feed-forward 3D reconstruction with 2D diffusion models to enable simultaneous 3D reconstruction and generation. Our geometry-aware leveling adapter, trained with a palette filtering strategy, effectively aligns diffusion priors with geometric control to fill missing regions and enhance novel views rendered from under-constrained 3D representations, while test-time masking refinement ensures artifact-free boundaries. Critically, unlike prior approaches, our refinement pipeline feeds enhanced views back into the reconstruction, leveling up the 3D representation. This unified framework achieves state-of-the-art performance on both novel-view synthesis and depth estimation on public datasets.

\bibliographystyle{splncs04}
\bibliography{main}
\clearpage
\setcounter{page}{1}
\maketitlesupplementary
% {
%     \newpage
%     \twocolumn[
%         \centering
%         \Large
%         \textbf{Supplementary Material}\\
%         \vspace{0.5em}
%         \vspace{1.0em}
%     ]
% }

\renewcommand{\thefigure}{S\arabic{figure}}
\setcounter{figure}{0}
\renewcommand{\thetable}{S\arabic{table}}
\setcounter{table}{0}
\renewcommand{\thesection}{\Alph{section}}
\setcounter{section}{0}

\section{Additional Implementation Details}

\noindent\textbf{Palette Filtering Strategy.}
To ensure dataset quality, we apply a palette-consistency filter defined as $\mathcal{S}_{\text{palette}} > \eta_{\text{palette}}$, where $\mathcal{S}_{\text{palette}}$ measures intensity distribution of the ground truth image, where pixels are inside the masked region (from rendering). Here we set $\eta_{\text{palette}} = 0.68$, following the proportion of samples within one standard deviation of the mean under a normal distribution. We also test other combinations of the standard and mean and test their performance, which is shown in Table~\ref{tab:pallete_std}. This threshold achieves the best trade-off: it removes visually inconsistent samples while retaining high-fidelity renderings, thereby significantly improving training stability and final perceptual quality without overly shrinking the dataset.
\\

\noindent\textbf{Mask Refinement with Morphological Operations.}
The refined mask is obtained through a two-stage morphological post-processing of the initial segmentation mask. 
We first apply closing with a $5 \times 5$ disk kernel ($\mathbf{K}_{5}$) to eliminate small holes and connects nearby regions. 
Subsequently, we perform dilation with a $20 \times 20$ dilation kernel ($\mathbf{K}_{20}$) to recover the main object contour while still suppressing noise. 
This asymmetric design compensates for the systematic under-segmentation tendency of the off-the-shelf segmenter on our data, yielding masks that are both clean and sufficiently tight for the refinement task.
\\

\noindent\textbf{Details for Diffusion Condition.}
We employ multiple conditioning signals in the diffusion model, including the coarse image, binary mask, text prompt, reference image, and geometry tokens. These conditions are injected at different stages of the pipeline following the setting in T2I~\cite{mou2024t2i}.
\\

\noindent\textbf{Coarse Renderings and Mask.}
The coarse image is first resized to $512 \times 512$ pixels and then encoded by the VAE encoder into a $64 \times 64$ latent representation (4 channels). 
The binary mask is converted to a single-channel grayscale image and resized to $64 \times 64$. 
After spatial alignment, the coarse image latent and the mask are concatenated channel-wise with the noised latent from the Stable Diffusion scheduler, forming the initial input to the Diffusion UNet.
\begin{equation}
\mathbf{Z}_{\text{input}} = \underbrace{\text{concat} \Big( \mathbf{Z}_t,\; \underbrace{\text{VAE}(\bar{I}),\; {M}^{refine}}_{\text{5 channels (condition)}} \Big)}_{\text{9 channels total}},
\end{equation}
\\

\noindent\textbf{Reference Image and Geometry Tokens.}
The reference image and geometry tokens are fed into the Leveling Adapter. 
They are first fused via a cross-attention module to produce a joint conditioning feature. 
This feature is subsequently processed by the adapter blocks to generate conditioning vectors that are injected into the Diffusion UNet through cross-attention layers.
\\

\noindent\textbf{Text Prompt.}
The text prompt guides the denoising process. We use \textit{"inpaint the image and remove degradation"} as our text prompt.
The text prompt is first encoded into a fixed-length embedding by the CLIP text encoder.
The resulting text embeddings are then used in the cross-attention layers of the Diffusion UNet to condition the generation at every denoising step.
\\

\section{Evaluation Details}
\noindent\textbf{Data Selection.} We choose to evaluate on the MipNeRF360~\cite{barron2022mip}, VRNeRF~\cite{xu2023vr}, TartanAir~\cite{wang2020tartanair}, and ScanNet~\cite{dai2017scannet} datasets since they are unseen in the training process of both~\cite{wang2025vggt, jiang2025anysplat} and our method. Similar to the data selection criterion of AnySplat~\cite{jiang2025anysplat}. 
For each dataset, we randomly select 5 distinct scenes. 
In each scene, we first determine an inter-image sparsity (sampling step) from \{5, 10, 20\} according to the characteristics of that scene. 
Using the chosen sparsity, we uniformly sample frames from the entire image sequence. 
From these sampled frames, we then randomly select two as training anchor views such that there are 4 to 6 frames between them. 
All frames lying strictly between these two anchors (under the same sparsity) are used as novel views for evaluation. 
Additionally, we include one extra frame immediately before the first anchor and one immediately after the last anchor within the range of step \{1, 4\}, both sampled with the same sparsity.
\\

\section{Additional Ablation on Palette Filtering}
\begin{table}[t]
  \centering
  \caption{ \textbf{Comparison of different palette threshold on VRNeRF~\cite{xu2023vr}.}}
  \resizebox{0.8\linewidth}{!}{
  \begin{tabular}{@{}l|ccccc@{}}
    \toprule
    range & PSNR  $\uparrow$ & SSIM $\uparrow$  & LPIPS $\downarrow$ & FID $\downarrow$ & Met3R $\downarrow$ \\
    \midrule
    Baseline~\cite{jiang2025anysplat}  & 15.892 & 0.532 & 0.354 & 153.39 & 0.0322 \\
    mean $\pm$ 0.5 std (0.38) & 17.984 & 0.586 & 0.334 & 161.89 & \underline{0.0254} \\
    mean $\pm$ 1 std (0.68) & \textbf{18.362} & \textbf{0.610} & \textbf{0.314} & \textbf{141.15} & \textbf{0.0242} \\
    
    mean $\pm$ 1.5 std (0.86) & \underline{18.106} & \underline{0.594} & \underline{0.324} & \underline{151.48} & {0.0266} \\
    \bottomrule
    \end{tabular}
  }
  \vspace{-2mm}
  \label{tab:pallete_std}
\end{table}
Table~\ref{tab:pallete_std} studies the effect of different palette thresholds on VRNeRF~\cite{xu2023vr}. We vary the concentration range used to compute the palette score, namely mean $\pm 0.5$ std, mean $\pm 1$ std, and mean $\pm 1.5$ std, which correspond to thresholds of 0.38, 0.68, and 0.86, respectively. Among these settings, mean $\pm 1$ std achieves the best overall performance, improving PSNR/SSIM to 18.362/0.610 and reducing LPIPS/FID/Met3R to 0.314/141.15/0.0242. In comparison, a narrower range is overly sensitive and yields inferior visual quality, while a wider range is too permissive and allows more low-information samples into training. These results indicate that the mean $\pm 1$ std criterion provides the best trade-off between filtering masked regions and preserving sufficiently diverse supervision.

\section{Additional Qualitative Results with Different Diffusion Condition Methods}

\begin{figure*}[t]
  \centering
  % \fbox{\rule{0pt}{1.5in} \rule{0.9\linewidth}{0pt}}
   \includegraphics[width=\linewidth]{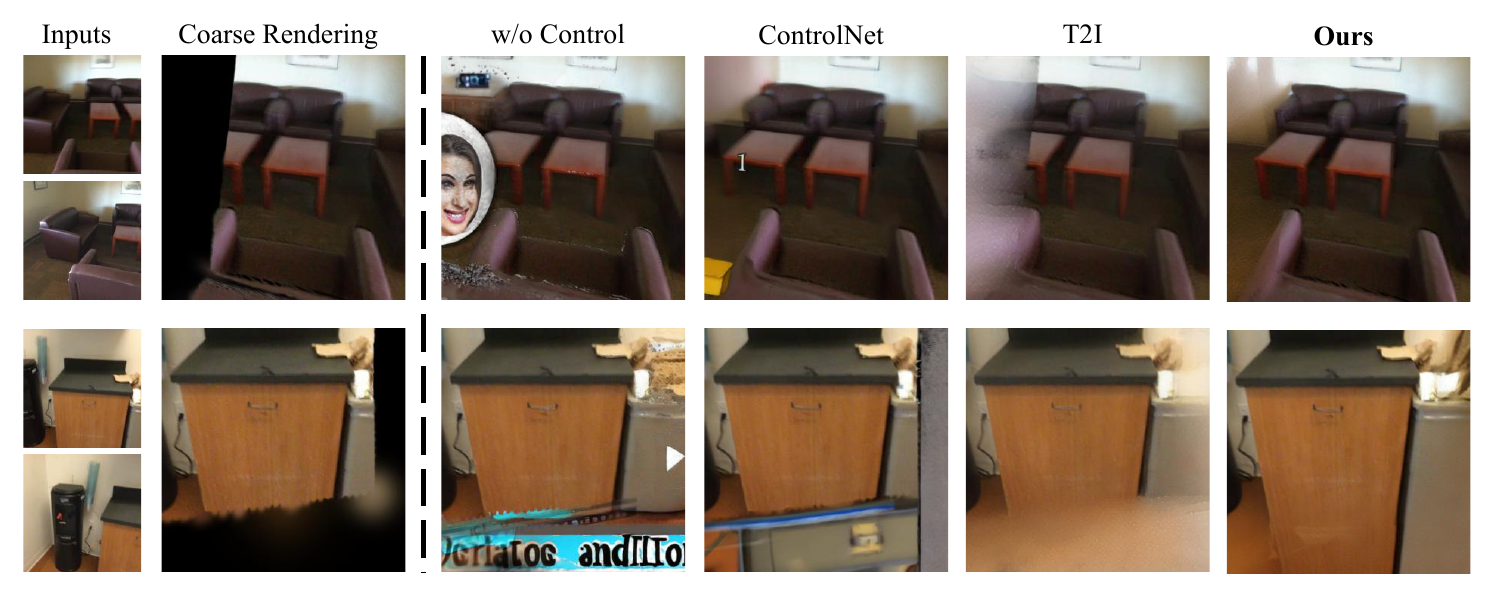}

   \caption{\textbf{Qualitative Results with Different Diffusion Condition Methods.} Our method outperforms other diffusion condition methods.}
   \label{fig:quali_diff}
\end{figure*}
In this section, we evaluate several diffusion condition methods and compare their qualitative performance in Fig.~\ref{fig:quali_diff}.
We utilize the standard Stable Diffusion 2~\cite{Rombach_2022_SD2} pipeline (image encoding, masked-region denoising, and latent merging). Stable Diffusion 2 (SD2) concatenates the noisy latent, the resized mask, and the masked original latents into a 9-channel UNet input for progressive denoising. The naive SD2 generally yields reasonable results but lacks explicit control conditions, often leading to implausible or artifact-prone outputs.
ControlNet and T2I-Adapter build upon the SD2 inpainting pipeline and inject a reference image as additional conditioning. While they provide stronger control over content and layout, they still struggle with precise geometric alignment and fine-grained pixel-level coherence. Our model fills this gap by generating 3D accurate extrapolated areas with geometry condition.

\begin{table*}[t]
\centering

\begin{minipage}[t]{0.48\textwidth}
  \centering
  \caption{\textbf{Quantitative comparison on nuScenes dataset}.  Our method demonstrates satisfactory generalization on driving scenes.}
  \resizebox{\linewidth}{!}{
  \begin{tabular}{@{}l|ccc}
    \toprule
    Method & PSNR$\uparrow$ & SSIM$\uparrow$ & LPIPS$\downarrow$ \\
    \midrule
    Baseline~\cite{jiang2025anysplat} & 15.21 & 0.463 & 0.474   \\
    
    Difix3D+~\cite{wu2025difix3d+} & \underline{15.49} & \underline{0.487} & \underline{0.450}  \\
    Ours & \textbf{17.79} & \textbf{0.583} & \textbf{0.413}  \\
    \bottomrule
  \end{tabular}
  }
    \label{tab:nuscene_tab}
\end{minipage}%
\hfill
\begin{minipage}[t]{0.48\textwidth}

  \centering
  \caption{\textbf{Quantitative comparison on SCARED~\cite{allan2021stereo_scared} dataset}. Our method shows satisfactory generalization results in the surgical domain.}
  \resizebox{\linewidth}{!}{
  \begin{tabular}{@{}l|ccc}
    \toprule
    Method & PSNR$\uparrow$ & SSIM$\uparrow$ & LPIPS$\downarrow$ \\
    \midrule
    Baseline~\cite{jiang2025anysplat} & 14.09 & 0.330 & \underline{0.410}   \\
    
    Difix3D+~\cite{wu2025difix3d+} & \underline{14.22} & \underline{0.355} & 0.430  \\
    Ours & \textbf{16.82} & \textbf{0.415} & \textbf{0.375}  \\
    \bottomrule
  \end{tabular}
  }
  
  \label{tab:scared_NVS}

  \end{minipage}

\end{table*}

\begin{table}[t]
  \centering
  
  \caption{ \textbf{Supplement depth estimation on SCARED~\cite{allan2021stereo_scared} dataset}. Our method achieves robust geometry-aware results.}
  \resizebox{0.7\linewidth}{!}{
  \begin{tabular}{@{}l|cccc@{}}
    \toprule
    Method & Abs Rel $\downarrow$ & RMSE $\downarrow$ & $\text{RMSE}_{log}$ $\downarrow$ & $\delta<1.25$ $\uparrow$  \\
    \midrule
    Baseline~\cite{jiang2025anysplat} & 1.192 & 20.25 & 3.700 & 0.195 \\
    Difix3D+~\cite{wu2025difix3d+}  & \underline{1.133}	& \underline{19.19} & \underline{0.824} & \underline{0.232} \\
    Ours& \textbf{1.117} & \textbf{19.05} & \textbf{0.818} & \textbf{0.234} \\
    \bottomrule
  \end{tabular}
  }
  % \vspace{-2mm}
  \label{tab:depth_sup}
\end{table}
\section{Additional Experiments on Driving Scenes}
\begin{figure}[t!]
  \centering
  % \fbox{\rule{0pt}{1.5in} \rule{0.9\linewidth}{0pt}}
   \includegraphics[width=\linewidth]{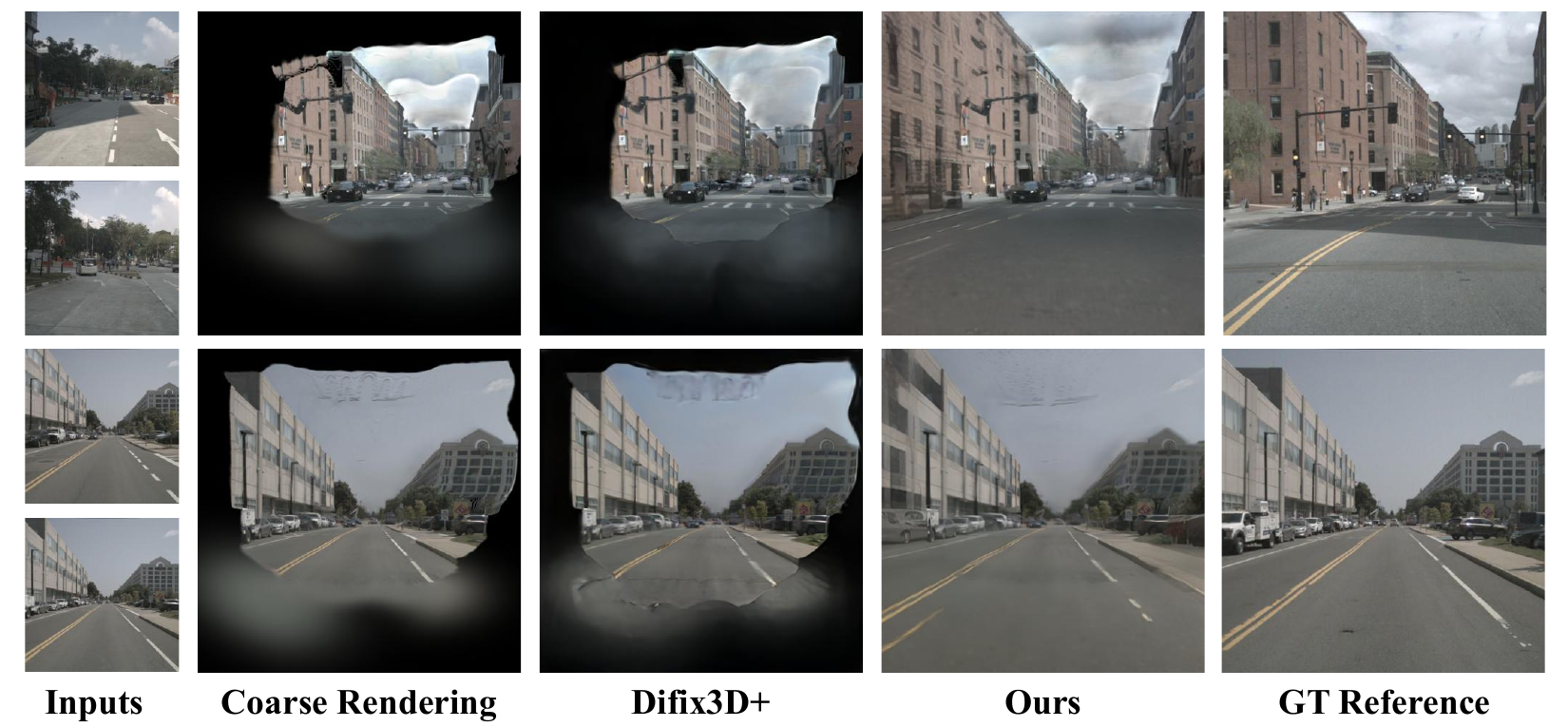}

   \caption{\textbf{Qualitative results on nuScenes dataset.} Our method outperforms other methods on novel view synthesis.}
   \label{fig:nuscenes_fig}
\end{figure}

To further demonstrate the generalization capability of Leveling3D, we directly apply our method, without any fine-tuning, to the challenging large-scale autonomous driving benchmark nuScenes~\cite{nuscenes2019}. As shown in Fig.~\ref{fig:nuscenes_fig} and Table~\ref{tab:nuscene_tab}, Leveling3D achieves satisfactory scene completion and refinement quality under extremely sparse input conditions (as few as 1–2 input images), substantially outperforming previous specialized methods specifically designed and trained on this dataset. These results highlight Leveling3D’s unique ability to generate complete and degradation-free 3D scenes from sparse input views. As world models are rapidly emerging as a core paradigm for embodied intelligence, our approach shows considerable potential as an effective and versatile component for future autonomous agents and general scene understanding systems.

\begin{figure*}[t!]
  \centering
  % \fbox{\rule{0pt}{1.5in} \rule{0.9\linewidth}{0pt}}
   \includegraphics[width=\linewidth]{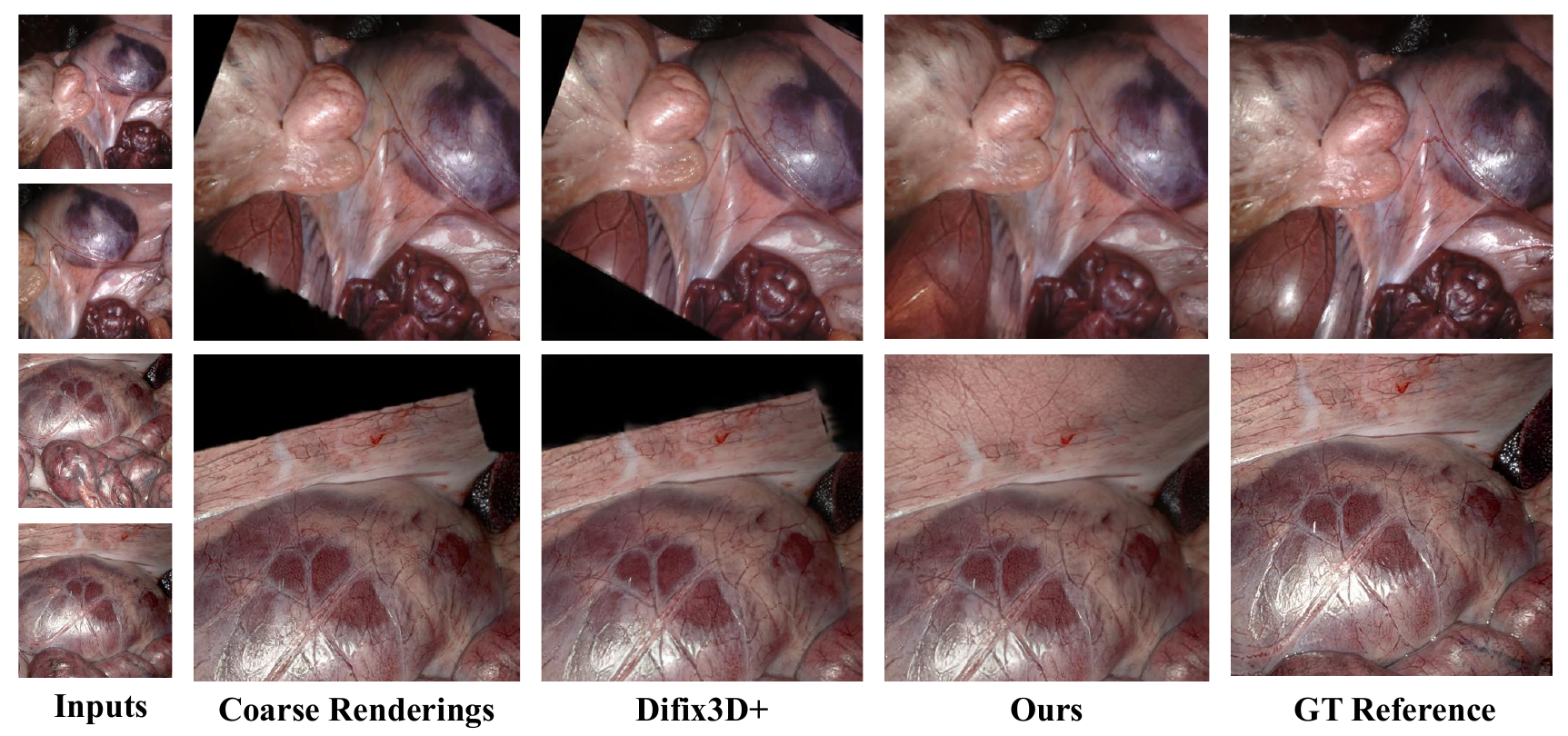}

   \caption{\textbf{Qualitative results on SCARED dataset.} Leveling3D achieves the best results for cross-domain surgical-scene novel-view synthesis and depth estimation, demonstrating robust generalization.}
   \label{fig:scared_NVS}
\end{figure*}

\section{Additional Experiments on Surgical Scenes}
We also validate the generalization of Leveling3D on even more challenging cross-domain SCARED~\cite{allan2021stereo_scared} dataset (Stereo Correspondence and Reconstruction of Endoscopic Data Challenge), a benchmark for surgical scene understanding. We assess both novel view synthesis and depth estimation under sparse input conditions. Quantitative results, reported in Table~\ref{tab:scared_NVS} for novel view synthesis and Table~\ref{tab:depth_sup} for depth estimation, demonstrate that Leveling3D achieves highly competitive performance compared to existing methods (see Fig.~\ref{fig:scared_NVS} for visual comparisons). These results highlight Leveling3D’s ability to effectively reconstruct surgical scenes by completing missing regions and reducing artifacts, even with limited input views and limited brightness. This capability underscores its significant potential for real-world surgical robotics applications, where acquiring input data is often challenging.

\section{Number of Parameters and Inference Time}

Table~\ref{tab:param_and_inference} reports the number of trainable parameters and inference time of baseline and all other diffusion control methods, as well as our method. The inference time is measured on the same device and averaged over 100 independent runs.
\\

\noindent\textbf{StableDiffusion2~\cite{Rombach_2022_SD2}.}
The Stable Diffusion 2 contains the VAE Model, CLIP Text Model, and UNet. Here we only finetune the parameters of UNet, which totally contains \textbf{865.83M} parameters. The averaged single inference time for StableDiffusion2 is \textbf{1.071s}.
\\

\begin{table}[t]
  \centering
  \caption{ \textbf{Comparison of number of parameters (Num. Param.) and inference time (Inf. Time).} * indicates diffusion model without control.}
  \resizebox{0.6\linewidth}{!}{
  \begin{tabular}{@{}l|cc@{}}
    \toprule
    Method & Num. Param.$\downarrow$ & Inf. Time$\downarrow$ \\
    \midrule
    % AnySplat~\cite{jiang2025anysplat} & - & 0.464s  \\
    % \midrule
    
    Naive Diffusion*~\cite{Rombach_2022_SD2} & 865.83M & {1.071s}  \\
    \midrule
    T2I~\cite{mou2024t2i} & \underline{83.51M} & \underline{1.186s} \\
    ControlNet~\cite{zhang2023ControlNet} & 369.34M & 1.548s  \\
    % Difix3D+~\cite{sauer2024SDturbo}  & 866.94M & 0.301s \\
    Ours & \textbf{68.74M} & \textbf{1.167s}  \\
    
    \bottomrule
  \end{tabular}
  }
  \vspace{-2mm}
  \label{tab:param_and_inference}
\end{table}

\noindent\textbf{T2I Adapter~\cite{mou2024t2i}.}
T2I Adapter training will freeze the Stable Diffusion components, focus mainly on the ResBlock-based T2I Adapter, and contain a total \textbf{83.51M} parameters. The averaged single inference time is \textbf{1.186s}.
\\

\noindent\textbf{Leveling3D.}
Our Leveling3D utilizes part of the T2I Adapter structure and adds a cross-attention module for the fusion of the geometry features and the reference images. We still freeze the Stable Diffusion components during the training, and the total number of trainable parameters is \textbf{68.74M} parameters. The averaged single inference time for our method is \textbf{1.167s}.
\\

\noindent\textbf{ControlNet~\cite{zhang2023ControlNet}.}
The ControlNet training also freezes the Stable Diffusion components, and the total number of training parameters in ControlNet is \textbf{369.34M}. The averaged single inference time is \textbf{1.548s}.

\section{Metrics}
\subsection{Image Metrics}

\noindent\textbf{LPIPS~\cite{johnson2016perceptual}}
We use the Learned Perceptual Image Patch Similarity (LPIPS) both as a training objective and as an evaluation metric. 
Unlike pixel-wise measures (e.g., MSE), LPIPS assesses perceptual similarity by computing distances between deep features from a pre-trained VGG network. 
It is defined as
\begin{equation}
\mathcal{L}_{\text{LPIPS}} = \frac{1}{L} \sum_{l=1}^{L} \alpha_l \Big\| \phi_l(\hat{I}) - \phi_l(I) \Big\|_2^2,
\end{equation}
where \(\phi_l(\cdot)\) denotes the feature map from the \(l\)-th layer of the network, and \(\alpha_l\) are learned weighting coefficients. 
Lower LPIPS scores indicate higher perceptual similarity.
\\

\noindent\textbf{PSNR}
Peak Signal-to-Noise Ratio (PSNR) measures pixel-level fidelity between generated and ground-truth images. 
It is defined as
\begin{equation}
\text{PSNR} = 10 \cdot \log_{10} \left( \frac{\text{MAX}^2}{\text{MSE}} \right),
\end{equation}
where \(\text{MAX}\) is the maximum possible pixel value (255 for 8-bit images) and \(\text{MSE}\) is the mean squared error. 
Higher PSNR values correspond to better reconstruction quality.
\\

\noindent\textbf{FID~\cite{heusel2017gans_FID}}
We employ the Fréchet Inception Distance (FID) to quantify the distributional similarity between generated and real images. 
FID computes the Wasserstein-2 distance between Gaussian distributions fitted to Inception-v3 feature embeddings:
\begin{equation}
\text{FID} = \|\mu_{\text{gen}} - \mu_{\text{gt}}\|_2^2 + \operatorname{Tr}\!\left( \Sigma_{\text{gen}} + \Sigma_{\text{gt}} - 2(\Sigma_{\text{gen}} \Sigma_{\text{gt}})^{1/2} \right),
\end{equation}
where \(\mu_{\text{gen}}, \mu_{\text{gt}}\) and \(\Sigma_{\text{gen}}, \Sigma_{\text{gt}}\) are the respective means and covariances of the feature distributions. 
Lower FID scores indicate better alignment with the real data distribution.
\\

\noindent\textbf{SSIM~\cite{wang2004image_ssim}}
Structural Similarity Index Measure (SSIM) evaluates perceptual quality by comparing luminance, contrast, and structural information. 
For images \(I_{\text{gen}}\) and \(I_{\text{gt}}\), it is defined as
\begin{equation}
\text{SSIM}(I_{\text{gen}}, I_{\text{gt}}) = \frac{(2\mu_{\text{gen}}\mu_{\text{gt}} + C_1)(2\sigma_{\text{gen,gt}} + C_2)}{(\mu_{\text{gen}}^2 + \mu_{\text{gt}}^2 + C_1)(\sigma_{\text{gen}}^2 + \sigma_{\text{gt}}^2 + C_2)},
\end{equation}
where \(\mu\) and \(\sigma^2\) are the mean and variance of pixel intensities, \(\sigma_{\text{gen,gt}}\) is the covariance, and \(C_1\), \(C_2\) are small constants for numerical stability. 
Higher SSIM values (closer to 1) reflect greater structural similarity.
\\

\subsection{Depth and Geometry Metrics}
We evaluate monocular depth estimation and 3D reconstruction quality using the following standard metrics.

\noindent\textbf{Abs Rel} \quad Absolute relative error measures the average relative difference:
\begin{equation}
\text{Abs Rel} = \frac{1}{N}\sum_{i=1}^{N} \frac{|d_{\text{pred}}^{(i)} - d_{\text{gt}}^{(i)}|}{d_{\text{gt}}^{(i)}},
\end{equation}
\\

\noindent\textbf{RMSE} \quad Root mean squared error quantifies absolute deviation in metric units:
\begin{equation}
\text{RMSE} = \sqrt{\frac{1}{N}\sum_{i=1}^{N} (d_{\text{pred}}^{(i)} - d_{\text{gt}}^{(i)})^2},
\end{equation}
\\

\noindent\textbf{$\text{RMSE}_{log}$} \quad Root mean squared error in log space, more robust to scale variations:
\begin{equation}
\text{RMSE}_{log} = \sqrt{\frac{1}{N}\sum_{i=1}^{N} \bigl(\log d_{\text{pred}}^{(i)} - \log d_{\text{gt}}^{(i)}\bigr)^2},
\end{equation}
\\

\noindent\textbf{{$\delta < 1.25$}} \quad Threshold accuracy: fraction of pixels where the ratio (or inverse) is within $1.25$:
\begin{equation}
\delta < 1.25 = \frac{1}{N}\sum_{i=1}^{N} 
\mathbf{1}\left(
\max\left(\frac{d_{\text{pred}}^{(i)}}{d_{\text{gt}}^{(i)}}, \frac{d_{\text{gt}}^{(i)}}{d_{\text{pred}}^{(i)}}\right) < 1.25
\right)
\end{equation}
Higher values are better for $\delta < 1.25$; lower values are better for the others. Here, $d_{\text{pred}}^{(i)}, \log d_{\text{gt}}^{(i)}$ are the corresponding pixel depths for the prediction and the ground truth, $N$ is the number of valid pixels used for depth evaluation.
\\

\noindent\textbf{Met3R~\cite{asim25met3r}} \quad The original Met3R score is used to compute the multi-view 3D consistency of generated images. A lower Met3R score $ \mathrm{MEt3R}(\cdot, \cdot)\in[0,2]$ indicates better image consistency. It is defined as:
\begin{equation}
\mathrm{MEt3R}(\mathbf{I}_1, \mathbf{I}_2) = 1 - \frac{1}{2} \Bigl( S(\mathbf{I}_1, \mathbf{I}_2) + S(\mathbf{I}_2, \mathbf{I}_1) \Bigr)
\end{equation}
where $S(\cdot, \cdot)$ means similarity calculation as in~\cite{asim25met3r}, and $\mathbf{I}$ means input image.
To extend this metric to image sequences, we first arrange the generated images in spatial order and then form consecutive pairs from the sequence, computing the average of their Met3R scores. This is defined as:
\begin{equation}
\mathrm{MEt3R}_{\text{seq}} \left( \{\mathbf{I}_i\}_{i=0}^{N-1} \right)
= \frac{1}{N-1} \sum_{i=0}^{N-2} \mathrm{MEt3R}(\mathbf{I}_i, \mathbf{I}_{i+1})
\end{equation}

\begin{figure*}[t]
  \centering
  % \fbox{\rule{0pt}{1.5in} \rule{0.9\linewidth}{0pt}}
   \includegraphics[width=0.9\linewidth]{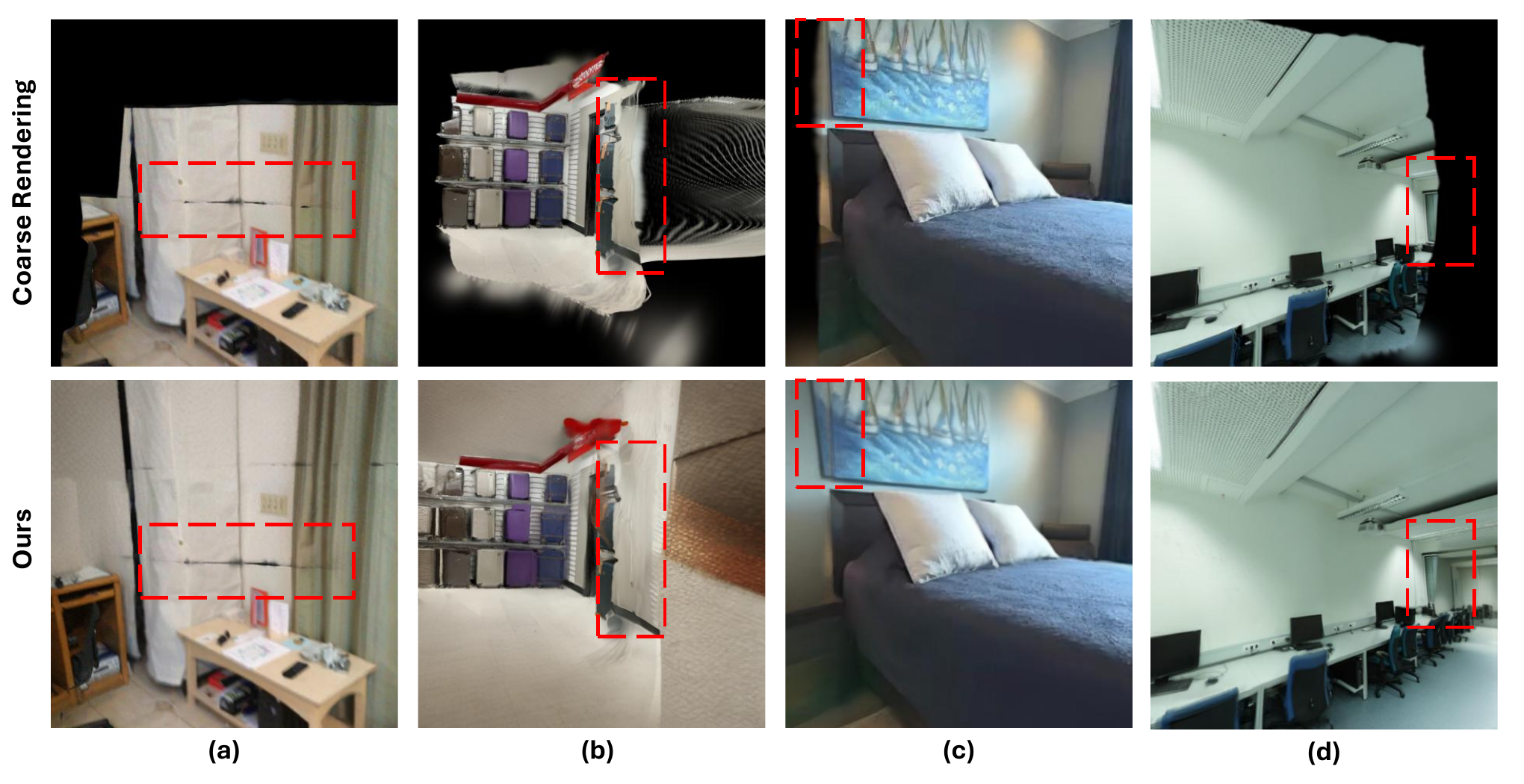}

   \caption{\textbf{Failure cases.} The highlighted areas are the key to failure.}
   \label{fig:failure_cases}
\end{figure*}

\section{Failure Cases Analysis}
Our method still exhibits failure cases, mainly due to two limitations. 
First, the generative model often generates artifacts when the target object is small or its reference features are overlooked. 
Second, the reconstruction baseline~\cite{jiang2025anysplat} suffers from issues in challenging regions~\cite{jiang2025anysplat} (e.g., textureless surfaces and large viewpoint changes), which directly impact the quality of our generated scenes.

We show representative failure cases in Fig.~\ref{fig:failure_cases}.
Cases (a) and (b) stem from inherent limitations of the baseline AnySplat initialization. In (a), AnySplat introduces unexpected artifacts (e.g., dark streaks) outside the mask, which our method cannot correct since they lie beyond the designated refinement region. In (b), severe degradation or collapse of the initial Gaussian primitives results in extremely noisy and unstructured representations, leading to rough and unreasonable novel views. Cases (c) and (d) highlight a current limitation of Leveling3D itself: when extrapolation boundaries or object terminations lie very close to the mask border, our model sometimes fails to detect that the structure has already been fully revealed. Consequently, it either slightly over-extends existing objects (c, continuing a mural beyond its true endpoint) or overlooks clear termination cues (d, ignoring an exposed room corner and erroneously extending the geometry farther than it should). These cases typically occur under extremely aggressive extrapolation settings and suggest promising directions for future improvements, such as boundary-aware regularization or explicit modeling of termination signals.

\end{document}